\documentclass[runningheads]{llncs}

\usepackage{eccv}

\usepackage{eccvabbrv}
\usepackage{wrapfig}
\usepackage{caption}
\usepackage{microtype}
\usepackage{graphicx}
\usepackage{booktabs}
\usepackage{indentfirst}
\usepackage[accsupp]{axessibility}  
\usepackage{multicol}

\usepackage[pagebackref,breaklinks,colorlinks,citecolor=eccvblue]{hyperref}

\usepackage{orcidlink}

\usepackage{xcolor}
\usepackage{overpic}
\usepackage{bm}
\usepackage{multicol}
\usepackage{multirow}
\usepackage{pifont}
\usepackage{color}
\usepackage{multirow}
\usepackage{overpic}
\usepackage{arydshln}
\usepackage{makecell}
\usepackage{epsfig}
\usepackage[T1]{fontenc}
\usepackage{hhline}
\usepackage{xcolor}
\usepackage{pifont}
\usepackage{enumitem}
\usepackage{colortbl}
\usepackage{verbatim}

\usepackage{verbatim}
\usepackage{amsmath}
\usepackage{amssymb}
\usepackage{booktabs}
\usepackage{pifont}

\definecolor{red}{rgb}{1,0,0}
\definecolor{blue}{RGB}{0,112,192}
\definecolor{purple}{RGB}{153,150,194}
\definecolor{green}{RGB}{0,255,0}
\definecolor{mygray}{RGB}{110,110,110}
\definecolor{mygray1}{RGB}{70,70,70}
\definecolor{dark_green}{RGB}{109,180,115}
\definecolor{dark_blue}{RGB}{0,0,246}
\definecolor{dark_purple}{RGB}{112,48,160}

\def\ie{\textit{i.e.}}

\begin{document}

\newcommand\blfootnote[1]{%
  \begingroup
  \renewcommand\thefootnote{}\footnote{#1}%
  \addtocounter{footnote}{-1}%
  \endgroup
}
\title{Bidirectional Progressive Transformer for Interaction Intention Anticipation} 


\author{Zichen Zhang\inst{1}\and
Hongchen Luo\inst{1} \and
Wei Zhai\inst{1, *} \and
Yang Cao\inst{1,2} \and
Yu Kang\inst{1,2}}

\authorrunning{Zhang et al.}

\institute{University of Science and Technology of China \and
Institute of Artificial Intelligence, Hefei Comprehensive National Science Center\\
\email{\{zhangzichen@mail., lhc12@mail., wzhai056@, kangduyu@, forrest@\}ustc.edu.cn}}

\maketitle
\begin{abstract}
\blfootnote{*Corresponding Author.}
Interaction intention anticipation aims to jointly predict future hand trajectories and interaction hotspots.
Existing research often treated trajectory forecasting and interaction hotspots prediction as separate tasks or solely considered the impact of trajectories on interaction hotspots, which led to the accumulation of prediction errors over time.
However, a deeper inherent connection exists between hand trajectories and interaction hotspots, which allows for continuous mutual correction between them. Building upon this relationship, a novel \textbf{\textit{B}}idirectional pr\textbf{\textit{O}}gressive \textbf{\textit{T}}ransformer (\textbf{\textit{BOT}}), which introduces a Bidirectional Progressive mechanism into the anticipation of interaction intention is established. Initially, \textbf{\textit{BOT}} maximizes the utilization of spatial information from the last observation frame through the Spatial-Temporal Reconstruction Module, mitigating conflicts arising from changes of view in first-person videos. Subsequently, based on two independent prediction branches, a Bidirectional Progressive Enhancement Module is introduced to mutually improve the prediction of hand trajectories and interaction hotspots over time to minimize error accumulation.
Finally, acknowledging the intrinsic randomness in human natural behavior, we employ a Trajectory Stochastic Unit and a C-VAE to introduce appropriate uncertainty to trajectories and interaction hotspots, respectively. 
Our method achieves state-of-the-art results on three benchmark datasets \textbf{\textit{Epic-Kitchens-100}}, \textbf{\textit{EGO4D}}, and \textbf{\textit{EGTEA Gaze+}}, demonstrating superior in complex scenarios. 

\end{abstract}

\section{Introduction}
\label{sec:intro}

The pursuit of future prediction on egocentric videos is centered on anticipating intentions and behaviors observed from a first-person perspective~\cite{rodin2021predicting}. 
Interaction intention can be reflected in the trajectory of hand motion and contact points on the next-active object~\cite{liu2022joint}. This task is novel but holds significant potential value, such as AR~\cite{AR1,AR2}, robot skill acqusition~\cite{RSS,geng2022end,bharadhwaj2023zero,mandlekar2020learning,bahl2022human}, hand reconstruction~\cite{grasp,grasp2} and pose estimation~\cite{pose_esti,pose2}.

\begin{figure}[tb]
\centering
\begin{overpic}[width=0.65\linewidth]{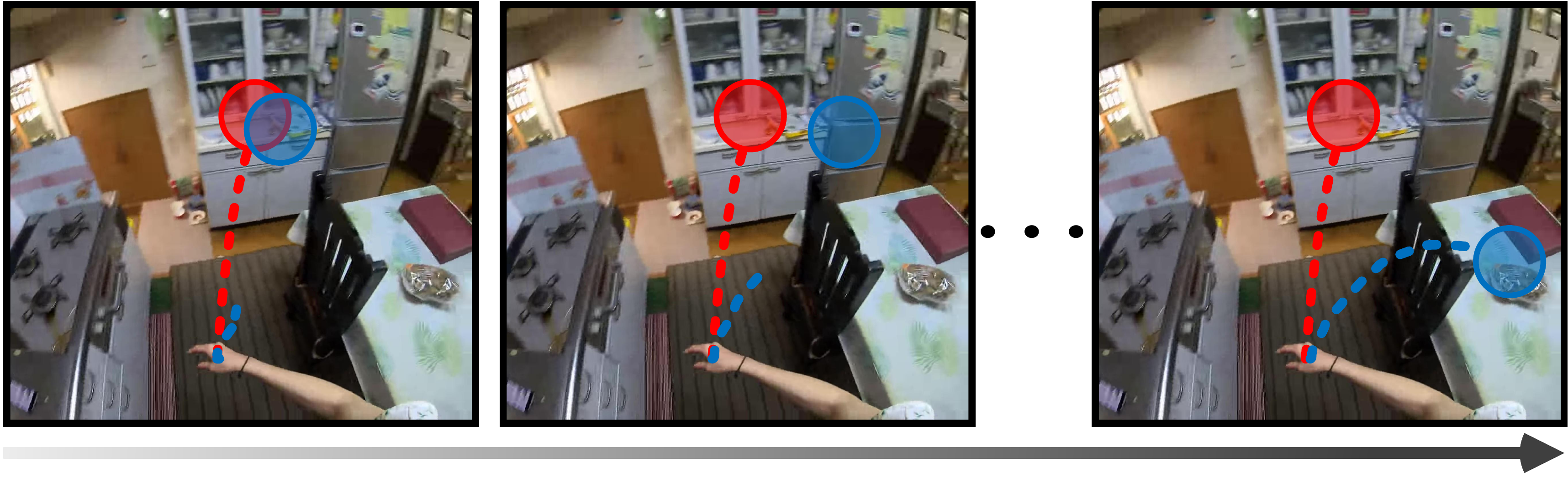}
\put(25,-2.){{\textbf{Error Accumulation Over Time}}}
\end{overpic}
\vspace{-2pt}
\caption{\textbf{Error accumulation} when only considering the influence from hand trajectories to interaction hotspots. \textcolor{red}{Red} ones represent the ground truth, while \textcolor{blue}{Blue} ones are predictions.}
\label{fig:intro_simple1}
\vspace{-3mm}
\end{figure}

Most existing relevant research treated hand trajectory forecasting and interaction hotspots prediction as independent tasks.
Bao \textit{et al}.~\cite{bao2023uncertainty} proposed to predict the 3D hand trajectory from a first-person view. However, They overlooked the guiding role of interactable objects on trajectories. The prediction process was incomplete, leading to significant errors at the end of trajectories.
On the other hand, there was extensive research on interaction hotspots~\cite{fang2018demo2vec_aff,kjellstrom2011visual,myers2015affordance,song2013predicting_aff,Luo_2022_CVPR,luo2023learning,hotspots}. Nevertheless, these studies often focused on the characteristics of the objects themselves, resulting in lower integration with subjective human behaviors in complex scenes and less accurate predictions. Other studies~\cite{liu2020forecasting,liu2022joint} jointly predicted future hand trajectories and interaction hotspots, improving their integration. However, these studies only considered the unidirectional influence of hand trajectories on contact points. Consequently, errors accumulated continuously with the prediction time steps (Fig.~\ref{fig:intro_simple1}).

\begin{figure}
\label{sec:intro}
\centering
\scriptsize
\begin{overpic}[width=0.60\linewidth]{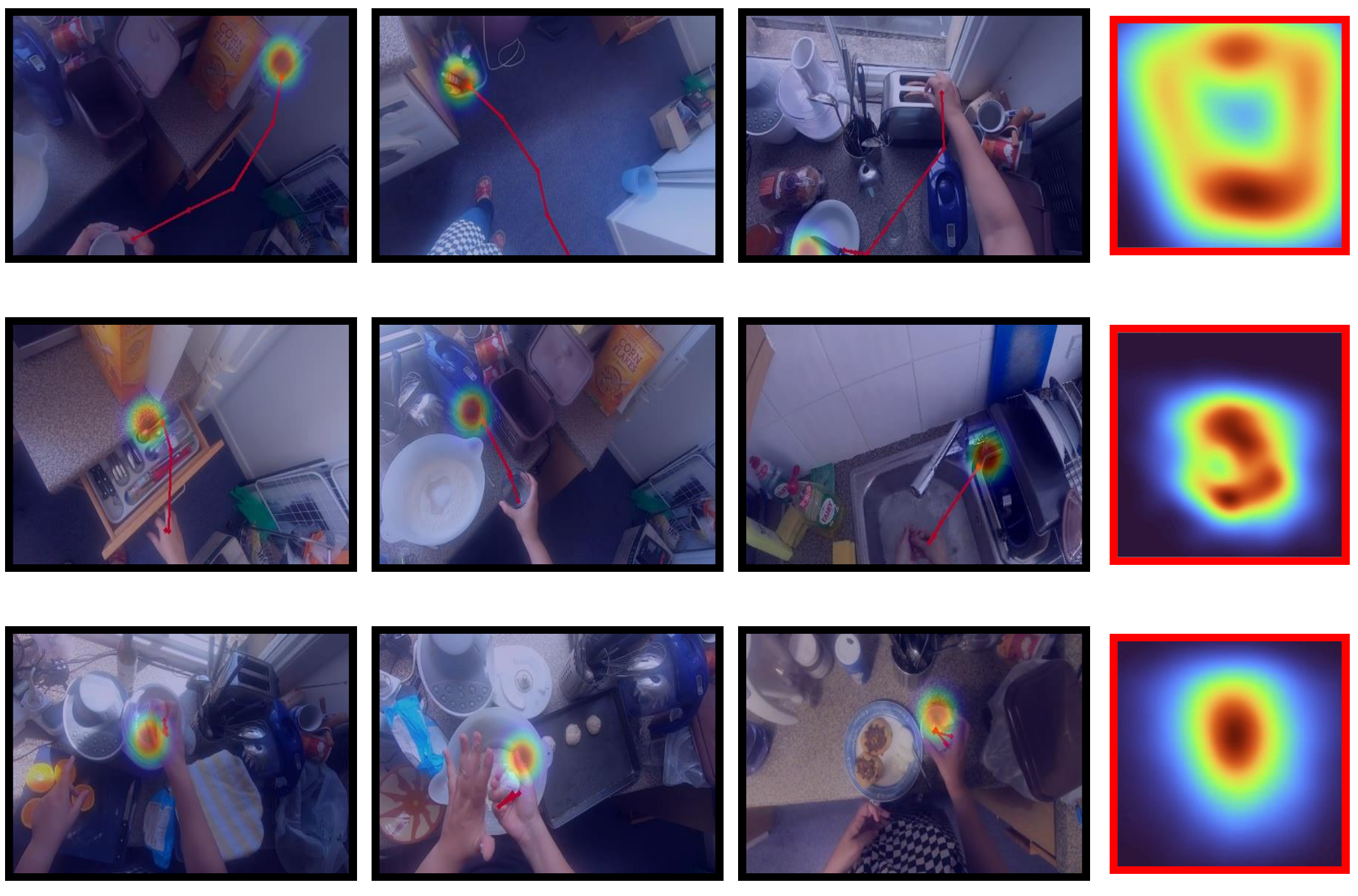}
\put(26,-2.2){{\textbf{(c) Short-length trajectories}}}
\put(23,21.2){{\textbf{(b) Medium-length trajectories}}}
\put(26,43.8){{\textbf{(a) Long-length trajectories}}}
\end{overpic}
\caption{\textbf{Inherent connection} between different hand trajectory categories and their corresponding contact points distribution. This is conducted on Epic-Kitchens-100~\cite{Damen2018EPICKITCHENS}.}
\label{fig:insight_2}
\end{figure}

In fact, this one-way causation was insufficient, as it disregarded the spatial correlation between the two factors. Conversely, there exists a mutual relationship between the predictions of hand trajectories and interaction hotspots.
On one hand, the derived directions of hand trajectories suggest the approximate areas where contact points occur. On the other hand, there is an inherent connection between the spatial distribution of contact points and hand trajectory patterns, as in Fig.~\ref{fig:insight_2}, which implies that examining the distribution of contact points can provide insights into the development of hand trajectories. Leveraging this relationship, to mitigate the process of error accumulation, we establish a \textbf{Bidirectional Progressive} (\textbf{Bi-Progressive}) mechanism, allowing continuous mutual refinement between trajectories and interaction hotspots throughout the entire prediction period, improving the accuracy of joint predictions.

In this paper, an innovative
 \textbf{B}idirectional pr\textbf{O}gressive \textbf{T}ransformer (\textbf{BOT}) is proposed, introducing the Bi-Progressive mechanism mentioned earlier between two independent predicting branches. 
Firstly, the raw video input undergoes a Spatial-Temporal Reconstruction Module, which integrates the spatial information from the last observation frame and the temporal information among all observation frames. This process mitigates spatial information conflicts arising from changes in the field of view in first-person videos.
Secondly, we establish separate dual branches for hand trajectory forecasting and interaction hotspots prediction, endowing them with the capability to independently perform individual prediction tasks, laying the foundation for introducing the Bi-Progressive mechanism. 
Subsequently, we introduce cross-attention blocks as the primary implementation of the Bi-Progressive mechanism. At each time step, the predictions of hand trajectories and interaction hotspots are alternately updated, addressing inherent errors in individual predictions. 
Finally, following the logic of natural human behaviors~\cite{uncetainty1,uncertainty2,uncertainty3}, we incorporate a Trajectory Stochastic Unit and a C-VAE for both hand trajectories and interaction hotspots, respectively, introducing appropriate uncertainties to the final prediction results.

We evaluate our approach on Epic-Kitchens-100~\cite{Damen2018EPICKITCHENS}, EGO4D~\cite{grauman2022ego4d} and EGTEA Gaze+~\cite{egtea} datasets. 
\textbf{BOT} demonstrates notable advantages in predicting both hand trajectories and interaction hotspots, achieving superior performance in anticipating interaction intentions.
In summary, our contributions are as follows:

\begin{itemize}
\item[$\bullet$] A novel \textbf{BOT} is introduced to jointly predict future hand trajectories and interaction hotspots in egocentric videos, attaining the optimal predictive performance with more lenient input conditions. 
\item[$\bullet$] We introduce a novel \textbf{Bidirectional Progressive} mechanism that enhances the coherence between hand trajectories and interaction hotspots, thereby improving the completeness of the interaction intention.
\item[$\bullet$] The experimental results demonstrate that our method achieves state-of-the-art results across three benchmark datasets \textbf{Epic-Kitchens-100}~\cite{Damen2018EPICKITCHENS}, \textbf{EGO4D}~\cite{grauman2022ego4d}, and \textbf{EGTEA Gaze+}~\cite{egtea}, demonstrating superior in complex scenarios.
\end{itemize}

\section{Related Work}
\textbf{Action anticipation in egocentric videos}. Anticipating actions in an egocentric vision is a challenging task that includes short-term and long-term action anticipation~\cite{rodin2021predicting,plizzari2023outlook}. Short-term action anticipation aims to predict the next action in the near future~\cite{furnari2018leveraging,miech2019leveraging,furnari2020rolling,rulstm,nvlaobananticipate}, while long-term action anticipation focuses on the sequence of subsequent actions performed by the camera wearer~\cite{liu2022hybrid,long_term2,long-term1,long_term3}. 
These endeavors contribute significantly to the comprehension of actions in first-person videos. 
However, concerning interaction intention, action anticipation still yields information of relatively high dimensionality.
Anticipating a semantic label does not furnish direct guidance for the robot, as it lacks specific instructions on task execution.
Comparatively, hand trajectories and interaction hotspots offer a more intuitive representation of interaction intention, constituting a lower-dimensional depiction.

\textbf{Human trajectory forecasting}.
The forecasting of human trajectories was extensively studied over the years~\cite{adeli2021tripod,mangalam2021goals,adeli2020socially,parsaeifard2021learning,Choi_2019_ICCV,chen2023unsupervised}.
However, most of them were based on third-person video datasets, falling short of capturing intentions related to human-object interactions. Since hands serve as the primary medium through which humans interact with the world, predicting hand trajectories in egocentric videos can aid embodied AI in comprehending human-object interaction intentions. 
The approaches in~\cite{liu2020forecasting,liu2022joint,bao2023uncertainty} addressed the issue of hand trajectory prediction in egocentric videos. However, they did not account for the influence of contact points with objects on trajectories. This limitation resulted in the continuous accumulation of prediction errors over time.

\textbf{Interaction hotspots prediction}.
Interaction hotspots prediction aims to locate where the hand-object interaction happens.
Some research aimed to predict the next interactive object~\cite{furnari2017next,bertasius2016first,bertasius2017unsupervised,fan2018forecasting,dessalene2021forecasting}, going further, some work anticipated specific interaction regions on objects~\cite{liu2020forecasting,luo2023learning,hotspots,liu2022joint},
yielding more precise information for interaction.
Differing from image input, our task aims to generate more precise predictions regarding the regions of contact points by utilizing egocentric video inputs. The incorporation of first-person videos is intended to focus on precisely localizing interaction hotspots for distinct human motions.
In this task, we leverage the predicted hand trajectories to improve the accuracy of interaction hotspots predictions.

\section{Method}
\textbf{Problem Setup.} Given an original input video $\mathcal{I}=\left \{ I_{1},I_{2},\cdots,I_{\mathrm{N} }     \right \} $, where $I_\mathrm{N} $ is the last observation frame, the task is predicting future hand trajectories $\mathcal{H} = \left \{ h_\mathrm{{N+1} },h_{\mathrm{N+2} },\cdots ,h_{\mathrm{N+T-1} }  \right \}  $ in future keyframes $\left \{{I_{\mathrm{N+1} },I_{\mathrm{N+2} },\cdots , I_{\mathrm{N+T-1} }}\right \}$ and the contact point $\mathcal{O}_\mathrm{C} $ in the contact frame $I_{N+T}$. We subsequently convert the contact point $\mathcal{O}_\mathrm{C} $ into a heatmap by centering a Gaussian distribution, generating interaction hotspots $\mathcal{O}_\mathrm{I} $.
To ensure the continuity of the trajectory as well as the consistency of the background, both $\mathcal{H}$ and $\mathcal{O}_\mathrm{I} $ are projected onto the last observation frame, which enhances the overall coherence of the predictions~\cite{liu2022joint}.

\begin{figure}[tb]
    \centering
    \includegraphics[width=1.\linewidth]{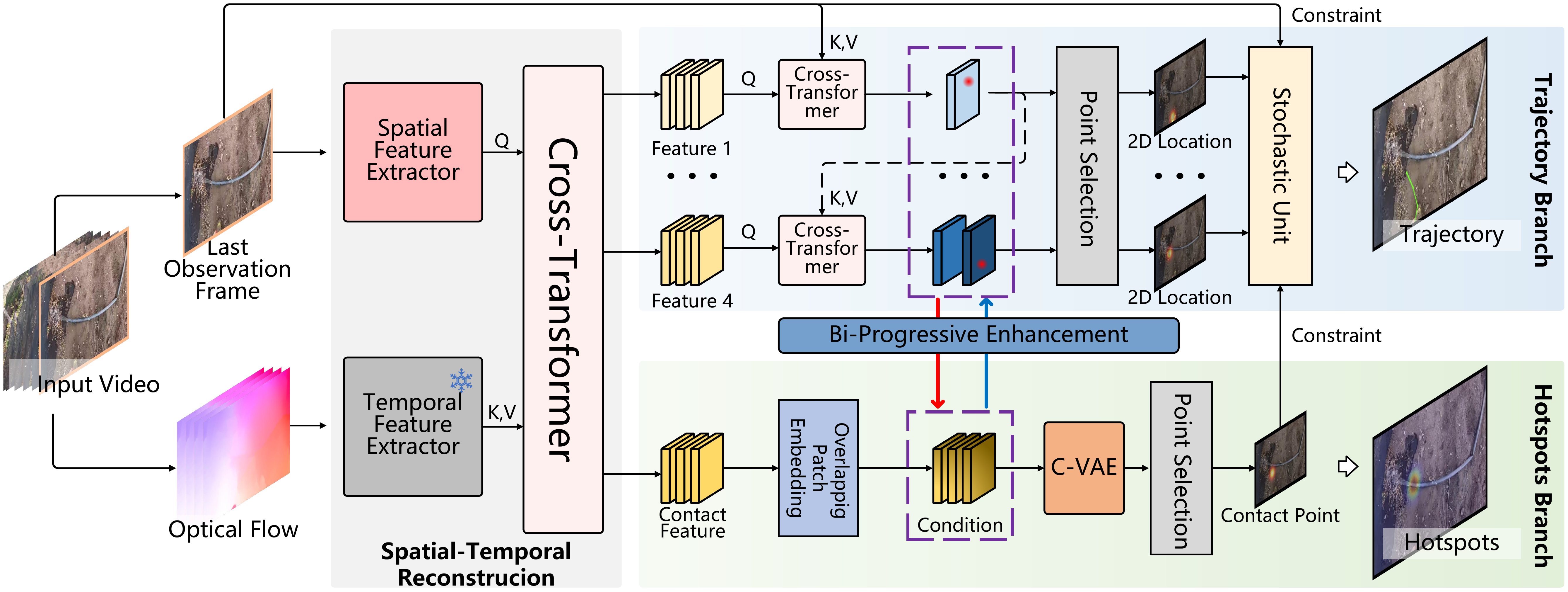}
    \put(-239,72){\rotatebox{-90}{\tiny{Eq.\ref{eq:1}-\ref{eq:recons}}}}
      \put(-171,97.6){\rotatebox{0}{\tiny{Eq.\ref{eq:4}}}}
      \put(-86,54.5){\rotatebox{0}{\tiny{Eq.\ref{eq:5}-\ref{eq:7}}}}
      \put(-55.5,66){\rotatebox{0}{\tiny{Eq.\ref{eq13}-\ref{eq14}}}}
      \put(-118,13){\rotatebox{0}{\tiny{Eq.\ref{eq:15}-\ref{eq:18}}}}
      \put(-172,8){\rotatebox{0}{\tiny{Eq.\ref{eq:extra}}}}
    \vspace{-6pt}
    \caption{\textbf{Overview of the Bidirectional Progressive Transformer}. It follows a dual-branch structure with a Bi-Progressive Enhancement Module between them to anticipate future hand trajectories and interaction hotspots.
    }
    \label{fig:method}
    \vspace{-3mm}
\end{figure}
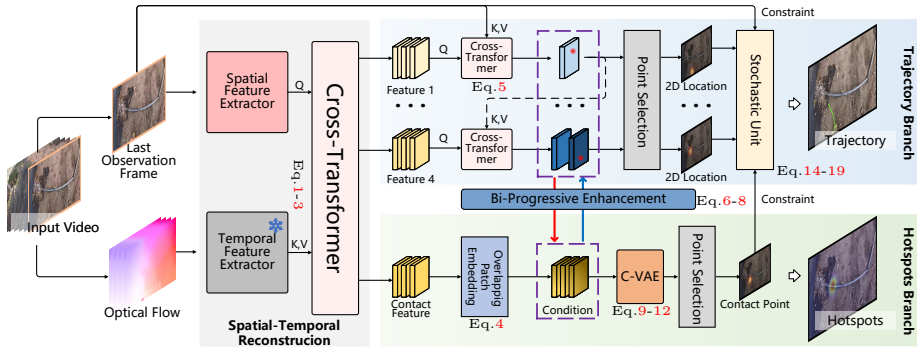

\subsection{Network}

\textbf{Overview.} As shown in Fig.~\ref{fig:method}, given the original input video $\mathcal{I}$, we aim to predict the future hand trajectory $\mathcal{H}$ and interaction hotspots $\mathcal{O}_\mathrm{I}$.
Initially, to alleviate information conflicts arising from changes in the field of view, the original input video undergoes a Spatial-Temporal Reconstruction Module that integrates spatial information from the last observation frame and temporal information from the entire video, producing spatial features for each future time step. Subsequently, these features are separated, with the last one serving as the contact feature while the preceding ones are for trajectory features. 
These features are then fed into the interaction hotspots and trajectory prediction branches, respectively. 
The former employs an overlap patch embedding layer to enhance attention at the center of the field of view, while the latter introduces temporal order to trajectory features through Cross-Transformers~\cite{chen2021crossvit}. 
Simultaneously, we employ a Bi-Progressive Enhancement Module consisting of cross-attention blocks between dual branches, introducing a Bi-Progressive mechanism to the joint prediction.
This alternately and iteratively corrects the features of trajectories and interaction hotspots in chronological order, thereby mitigating the accumulation of prediction errors over time. 
Finally, considering the inherent uncertainty in natural human behaviors~\cite{uncetainty1,uncertainty2,uncertainty3}, we introduce a Stochastic Unit for trajectory forecasting while employing a C-VAE \cite{sohn2015learning} for interaction hotspots, adding appropriate uncertainty during the completion of position decoding to achieve comprehensive predictions.

\textbf{Spatial-Temporal Reconstruction.} 
When dealing with egocentric videos, both 3D convolutions \cite{feichtenhofer2019slowfast} and self-attention mechanisms \cite{self-attention} 
struggle to address spatial information conflicts arising from changes in views. 
Moreover, predictions are projected onto the last observation frame.
Taking these factors into account, we establish the Spatial-Temporal Reconstruction Module.
On one hand, we employ optical flow containing more comprehensive motion information as the temporal input, on the other hand,
we isolate the last observation frame as the spatial input.
Subsequently, the temporal input is processed by a pre-trained Temporal Segment Network~\cite{wang2016temporal,rulstm}, extracting low-dimensional temporal features $\mathbf{X}$, and
the spatial input is passed through a Segformer-b2~\cite{xie2021segformer}, extracting spatial features $\mathbf{S}$ from the last observation frame.
$\mathbf{X}$ and $\mathbf{S}$ are then fed into a Cross-Transformer~\cite{chen2021crossvit}. The $i$-th block is expressed as:
\begin{align}
    \mathbf{ s}_{0} = \mathbf{s} _{patch} + \mathbf{s} _{pos}, \quad    \mathbf{x}  = \mathbf{x}  + \mathbf{x} _{pos} , \label{eq:1} \\
   \  { \mathbf{y} _{i} = \mathbf{s} _{i-1} + {\mathbf{MCA}  }  (\mathbf{LN} (\mathbf{s} _{i-1}),\mathbf{LN} (\mathbf{x} ))} ,   \\
    {\mathbf{s} _{i} = \mathbf{y} _{i} + \mathbf{FFN} (\mathbf{LN} (\mathbf{y} _{i}))}. \ \ \ \ \ \  
    \label{eq:recons}
\end{align}
\noindent Where $\mathbf{s}_{0}$ and $\mathbf{x}$ represent the original spatial and temporal inputs, respectively.
$\mathbf{s}_{patch}$ is encoded by overlap patch embedding layers from spatial features $\mathbf{S}$, enhancing the network's focus on the center of the field of view.
$\mathbf{x}$ is encoded by positional embedding layers from temporal features $\mathbf{X}$, improving the overall sequentiality.
Furthermore, $\mathbf{FFN}$ refers to the Feed Forward Layer, comprising a two-layer \textbf{MLP} with an expanding ratio $r$, and Layer normalization ($\mathbf{LN}$) is applied before each block. 
$\mathbf{MCA}$ stands for Multi-head Cross-Attention, where query, key, and value are processed through distinct linear layers: ${\mathbf{MCA} (\mathrm{X,Y} ) = {Attention}(\mathbf{W} ^{Q} \mathrm{X} ,\mathbf{W} ^{{K} } \mathrm{Y},\mathbf{W} ^{V} \mathrm{Y} )} $.
The output of this module corresponds to the spatial features for each prediction frame.

\textbf{Interaction Hotspots Prediction Branch.}
Due to the spatial distribution of contact points closer to the central field of view, the contact feature $\mathbf{F}_\mathrm{C}$ undergoes processing through an overlap patch embedding layer:
\begin{align}
    \mathbf{F}^\mathrm{o} _{\mathrm{C} } = overlap(\mathbf{F} _{\mathrm{C} }).
    \label{eq:extra}
\end{align}
Where $\mathbf{F}^\mathrm{o}_\mathrm{C}$ represents the processed contact feature.

\textbf{Hand Trajectory Prediction Branch.}
We employ original trajectory features as primary features and utilize features from the previously predicted frame as conditional features, subsequently merging them through the cross-attention block $\mathit{f}$, ensuring the coherence of the trajectory over time.
Assuming the trajectory features are  $\left \{{\mathbf{F}_{\mathrm{N+1} },\mathbf{F}_{\mathrm{N} +2},\cdots , \mathbf{F}_{\mathrm{N+T-1} }}\right \}$ and the feature map of the last observation frame is $\mathbf{F} _{\mathrm{N}}$,
they are updated in chronological order:
\begin{equation}
    {\mathbf{{F} }^\mathrm {o} _{{t} }}  = \mathit{f} (\mathbf{F} _\mathrm{t},\mathbf{F}_{\mathrm{t-1}}), \quad {t} \in [\mathrm{N+1,N+T-1} ].
    \label{eq:4}
\end{equation}
Where ${\mathbf{{F} }^\mathrm {o} _{{t} }}$ stands for the updated trajectory feature,
$t$ is the prediction time step.

\textbf{H-O Bi-Progressive Enhancement.}
Expanding on the groundwork laid by
two distinct prediction branches, and to mitigate the accumulation of errors as predictions unfold over time, we introduce a Bi-Progressive Enhancement Module (Fig.~\ref{fig:net2}), which alternately corrects the predictions of trajectories $\mathcal{H}$ and interaction hotspots $\mathcal{C}_\mathrm{I}$ at each time step $t$. 
Specifically, owing to observation frames and the continuity of hand trajectories, the trajectory feature $\mathbf{F}^\mathrm {o}_\mathrm{N+1}$ at the initial predicted time step $\mathrm{N+1}$ demonstrates a high level of accuracy. Exploiting this, we enhance the precision of the contact feature $\mathbf{F}^\mathrm {o}_\mathrm{C}$ through the cross-attention block $\textit{f}$, thereby minimizing the error with the ground truth. 
Moving to the subsequent time step $\mathrm{N+2}$, we employ the first refined contact feature $\mathbf{F}^\mathrm {r_1}_\mathrm{C}$ from the previous time step $\mathrm{N+1}$ as a condition to correct the trajectory feature $\mathbf{F}^\mathrm {o}_\mathrm{N+2}$.
Following this, the corrected trajectory feature is utilized to further refine the contact feature. This iterative correction process persists until all trajectory features have undergone a complete round of updates. It can be represented as a recursive procedure:
\begin{align}
    \mathbf{F }^\mathrm{r} _{\mathrm{N+1} } =  \mathbf{F}^\mathrm o _{\mathrm{N+1} },
    \ \ \ \ \ \ \ \ \ \ \ \ \ \ \ \ \ \ \ \ \label{eq:5} \\
    \mathbf{F}^\mathrm {r_1} _{\mathrm{C} } = \mathit{f}  ( \mathbf{F}^\mathrm {o} _{\mathrm{C} },\mathbf{F}^\mathrm{r} _{\mathrm{N+1} }), \ \mathbf{F} ^{\mathrm{r} }  _{\mathrm{N+2} } =  \mathit{f}(\mathbf{F} ^{\mathrm{o} }_{\mathrm{N+2} } ,\mathbf{F} ^{\mathrm{r_1} } _{\mathrm{C} } ), \label{eq:6}\\
    \cdots \nonumber \ \ \ \ \ \ \ \ \ \ \ \ \ \ \  \ \ \ \ \ \ \ \ \ \ \ \
    \\
    \mathbf{F} ^{\mathrm{r_3} } _{\mathrm{C} } = \mathit{f}  ( \mathbf{F} ^{\mathrm{r_2} } _{\mathrm{C} },\mathbf{F} ^{\mathrm{r} } _{\mathrm{N+3} }),  \ \mathbf{F} ^{\mathrm{r}} _{\mathrm{N+4} } =  \mathit{f}(\mathbf{F} ^{\mathrm{o} } _{\mathrm{N+4} } ,\mathbf{F} ^{\mathrm{r_3} }_{\mathrm{C} } ).
\label{eq:7}
\end{align}
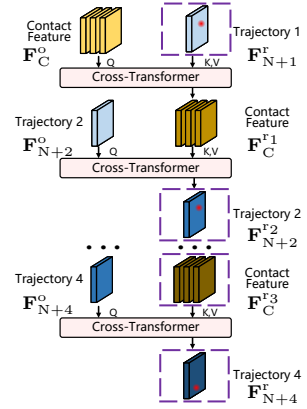
\begin{wrapfigure}{r}{0.32\textwidth}
\centering
\footnotesize
    \vspace{-6mm}
    \centering
    \includegraphics[width=0.99\linewidth]{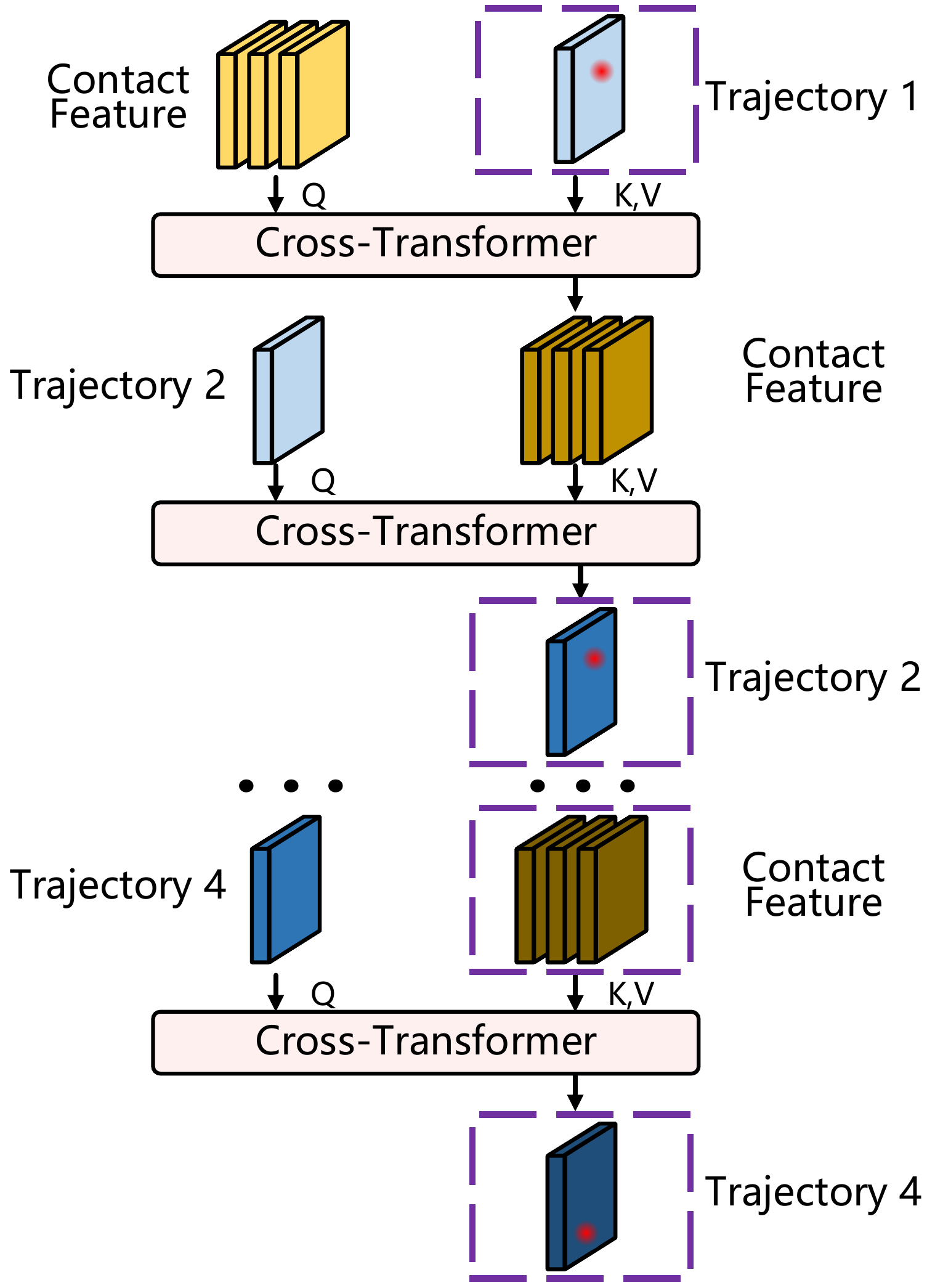}
    \put(-21,130){\rotatebox{0}{\scriptsize{$\mathbf{F}^\mathrm {r}_\mathrm{N+1}$}}}
    \put(-106,130){\rotatebox{0}{\scriptsize{$\mathbf{F}^\mathrm {o}_\mathrm{C}$}}}
    \put(-21,96){\rotatebox{0}{\scriptsize{$\mathbf{F}^\mathrm {r_1}_\mathrm{C}$}}}
    \put(-106,96){\rotatebox{0}{\scriptsize{$\mathbf{F}^\mathrm {o}_\mathrm{N+2}$}}}
    \put(-21,62){\rotatebox{0}{\scriptsize{$\mathbf{F}^\mathrm {r_2}_\mathrm{N+2}$}}}
    \put(-106,36){\rotatebox{0}{\scriptsize{$\mathbf{F}^\mathrm {o}_\mathrm{N+4}$}}}
    \put(-21,36){\rotatebox{0}{\scriptsize{$\mathbf{F}^\mathrm {r_3}_\mathrm{C}$}}}
    \put(-21,2.5){\rotatebox{0}{\scriptsize{$\mathbf{F}^\mathrm {r}_\mathrm{N+4}$}}}
    \vspace{-1mm}
    \caption{\textbf{Bi-Progressive Enhancement Module.} The output feature map is enclosed in \textcolor{dark_purple}{purple} boxes.}
    \label{fig:net2}
    \vspace{-8mm}
\end{wrapfigure}
Where $\mathbf{F}^{\mathrm{r}}_{t}$ represents the refined trajectory feature map at time step $t\in [\mathrm{N+1,N+T-1}]$, $\mathbf{F}^{\mathrm{r}_i}_\mathrm{C}$ denotes the contact feature after the $i$-th refinement.
Via this module, we attain refined features for both trajectories and interaction hotspots, significantly reducing error accumulation in predictions.

\textbf{Interaction Hotspots Decoder.}
Given the presence of inherent uncertainty associated with interaction hotspots~\cite{liu2022joint,Luo_2022_CVPR,luo2023learning}, 
we employ a C-VAE \cite{sohn2015learning} as the decoder.
Due to the excessive randomness associated with directly decoding the coordinates of contact points, we instruct the C-VAE to output a feature map and select the location of the maximum value from this map as the output, thereby confining the uncertainty within a reasonable range.
The C-VAE follows an encoder-decoder structure.
For the encoder block, we concatenate the input $\Gamma$ with the condition $\mathbf{C}$ through an encoding function $\mathbf{P}_\mathrm{e}$, obtaining features in latent space $\mathbf{Z}$ which parameterized by mean $\mu$ and co-variance $\sigma$.
In this context, the input $\Gamma$ is the ground truth heatmap, while the condition $\mathbf{C}$ is the refined contact feature $\mathbf{F}^{\mathrm{r_3} } _{\mathrm{C} } $.
In the decoder block, the first step involves sampling $\mathbf{Z}$ from the latent space and concatenating it with condition $\mathbf{C}$. 
Following that, the output heatmap $\hat{\Gamma}$ is reconstructed by a decoding function $\mathbf{P}_{d}$.
In summary, the entire process can be formulated as follows:
\begin{equation}
    \mu,\sigma = \mathbf {P}_{\mathrm{e} }({\Gamma },\mathrm{C}) , \quad \hat{{\Gamma }} = \mathbf{P}_{\mathrm{d} }(\mathrm{Z},\mathrm{C}),\quad \mathrm{Z} \sim \mathcal{N}(\mu,\sigma).
    \label{eq:15}
\end{equation}
Where $\mathcal{N}$ represents a normal distribution, $\hat{\Gamma}$ is the reconstructed heatmap.
Both encoding and decoding blocks consist of two \textbf{MLP} layers.
We introduce KL-Divergence and the reconstruction error as the overall loss:
\begin{align}
    \mathcal{L}_{recon}({\Gamma },\hat{\Gamma} )  =(\Gamma -\hat{\Gamma })^{2}, \ \ \ \ \ \ \ \ \ \\
    \mathcal{L}_{kl}(\mu,\sigma)  =-\mathbf{KL} \left [ \mathcal{N}(\mu,\sigma)^{2}\left |  \right | \mathcal{N}(0,1)   \right ], \\
    \mathcal{L}_o =  \mathcal{L}_{recon} +\lambda  \mathcal{L}_{kl}. \ \ \ \ \ \ \ \ \ \ \ \ 
    \label{eq:18}
\end{align}
In this paper, we set $\lambda = 0.01$.

\textbf{Hand Trajectory Decoder.}
Similar to interaction hotspots, hand trajectories also exhibit uncertainty~\cite{mangalam2020not,gupta2018social,uncertain10}. Given that trajectory feature maps with distinct distribution characteristics have been obtained through the Bi-Progressive Enhancement Module, in conjunction with sampling efficiency, we don't utilize C-VAE, rather, we introduce a Trajectory Stochastic Unit to incorporate uncertainty into trajectory forecasting by calculating uncertain regions for each prediction time step instead.
Since optimizing directly using point coordinates for loss calculation can pose challenges, we initially introduce a Gaussian distribution for each ground truth hand position to generate the ground truth heatmap. Subsequently, we calculate the loss between the corrected feature map and the generated ground truth heatmap, thereby enabling the network to learn the distribution of potential hand positions.
The associated loss function $\mathcal{L}_{h}$ is based on MSE Loss:
\begin{equation}
    \mathcal{L}_{h} = \frac{1}{\mathrm{T} -1}\sum_{t=\mathrm{N} +1}^{\mathrm{N+T} -1}(\mathbf{F}^\mathrm{r} _{t} - \mathbf{G} _{t} )^{2}.
\end{equation}
Where $ \mathbf{G} _{t}$ represents the ground truth heatmap at time step $t$. 
We designate the coordinates of the maximum value within each feature map as the predicted trajectory points, which are then fed into the stochastic unit to generate uncertain regions.
The size of the region is determined by the time step as well as the length of trajectory segments in preceding and subsequent stages. The shape is constrained by the original image dimensions and the orientation of the trajectory segments.
\begin{figure}[t]
    \centering
    \includegraphics[width=0.80\linewidth]{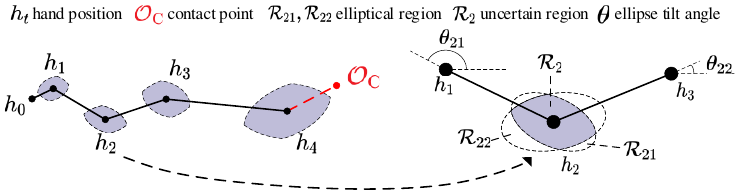}
    \caption{\textbf{Uncertain regions of the hand trajectory}. It is derived through the intersection of two ellipses, corresponding to the Eq.~\ref{eq13}-\ref{eq14}. (\ie, the \textcolor{purple}{purple} region).}
    \label{fig:math}
    \vspace{-2mm}
\end{figure}
Formally, the uncertain region (Fig.~\ref{fig:math}) can be shown as:
\begin{align}
\label{eq13}
    \mathcal{R}_{t} \sim (t,{h}_{t-1},{h}_{t},{h}_{t+1},\mathrm{W},\mathrm{L}).
\end{align}
Where $\mathcal{R}_{t}$ is the uncertain region in future time step $t$, 
$\mathrm{W}$ and $\mathrm{L}$ are the width and length of the input video, respectively.
$\mathcal{R}_{t}$ satisfies the following constraints:
\begin{equation}
\label{eq9}
    \mathcal{R} _{t} = \mathcal{R} _{t1} \cap \mathcal{R} _{t2}.
\end{equation}
Where $\mathcal{R}_{t1}$ and $\mathcal{R}_{t2}$ are elliptical regions constrained by $h_{t-1}$ and $h_{t+1}$, which can be represented as:
\begin{align}
    \frac{  ( \mathbf{x} _{t}\cos  \theta_{t\omega } -\mathbf{y} _{t}\sin \theta_{t\omega}   )^{2}  }{\mathbf{a} _{t\omega }^{2} }+\frac{   ( \mathbf{x} _{t}\sin \theta_{t\omega} +\mathbf{y} _{t}\cos \theta_{t\omega}  )^{2}  }{\mathbf{b} _{t\omega }^{2} }\le 1, \\
    \mathbf{a} _{t\omega  }=\alpha{\beta^{ t-1} \mathbf{d} ^{\frac{1}{2}  } ( h_{t+\omega -1},h_{t+\omega -2}   )  } , \          \mathbf{b} _{t\omega } = \sqrt{\frac{\mathrm{L}}{\mathrm{W}}} \mathbf{a} _{t\omega },  \ \ \  \label{eq:16}\\
    \theta_{t\omega  }= \tan^{-1} (\mathbf{k} (h_{t+\omega-1},h_{t+\omega-2})), \ \ \ \ \ \ \ \ \ \ \ \ \ \ \ \\
    \omega \in  [1,2], \ t\in [1,2,3,4], \ \mathcal{O}_\mathrm{C} = h_5. \ \ \ \  \ \ \ \ \ \ \ \ \ \
    \label{eq14}
\end{align}

\noindent Specifically, $\alpha$ denotes the overall scale of $\mathcal{R}_{t}$, $\beta$ signifies the temporal growth rate, $\theta$ stands for the ellipse tilt angle, $(\mathbf{x} _t,\mathbf{y} _t)$ represent the coordinates of $h_t$. Additionally, $\mathbf{k}\mathrm{(X,Y)} $ and $\mathbf{d}\mathrm{(X,Y)} $ represent the slope and Euclidean distance between points $\mathrm{X} $ and $\mathrm{Y} $, respectively.
In \textbf{BOT}, we set $\alpha=1.5$ and $\beta=2$.

\subsection{Training and Inference}

\textbf{Training.}
During training, we freeze parameters in TSN~\cite{wang2016temporal}.
Training loss combines both trajectory loss and interaction hotspots loss:
\begin{equation}
    \mathcal{L} = \mathcal{L}_{h} + \varphi  \mathcal{L}_{o}.
\end{equation}
In this paper, we set $\varphi  = 1$.

\textbf{Inference.}
We use a Trajectory Stochastic Unit and a C-VAE to introduce uncertainty during inference.
Following previous work~\cite{mangalam2020not,liu2022joint}, we report the minimum among 20 samples for trajectory evaluation.

\section{Experiments}

\subsection{Datasets}
We conducted experiments on three first-person video datasets 
\textbf{Epic-Kitchen
s-100}~\cite{Damen2018EPICKITCHENS}, \textbf{EGO4D}~\cite{grauman2022ego4d}, and \textbf{EGTEA Gaze+}~\cite{egtea}. Epic-Kitchens-100~\cite{Damen2018EPICKITCHENS} and EGTEA Gaze+~\cite{egtea} predominantly include kitchen scenes, while EGO4D~\cite{grauman2022ego4d} encompasses a broader range of activities such as handicrafts, painting, and outdoor tasks. For each dataset, the observation time is set to 2 seconds. We configure the prediction time horizon $\mathrm{T}$ to 5, corresponding to 4 trajectory points and one contact point for each hand. Specifically, we predict 1-second trajectories on Epic-Kitchens-100~\cite{Damen2018EPICKITCHENS} and EGO4D~\cite{grauman2022ego4d}, while on EGTEA Gaze+ \cite{egtea}, 0.5-second trajectories are forecast. Additional dataset information is available in the supplementary materials.

\subsection{Implementation Details}
In EGO4D~\cite{grauman2022ego4d}, We use the method~\cite{liu2022joint} to automatically generate hand trajectories, while contact points are manually annotated based on the last observation frame and the contact frame.
We employ GMFLOW~\cite{xu2022gmflow} to generate optical flow from EGO4D~\cite{grauman2022ego4d} and EGTEA Gaze+~\cite{egtea}.
For each dataset, we resize the original videos to 456 $\times$ 256 size, extracting both trajectory and contact features of size 114 $\times$ 64.
In the Spatial-Temporal Reconstruction Module, we set the dimension $D$ in Cross-Transformer to 1024, with depth $= 6$ and attention head number $= 8$.
For the trajectory forecasting branch and H-O Bi-Progressive Enhancement Module, we configure the dimension $D$ to 256, with depth $= 2$ and head number $= 8$.
Additionally, for the introduced C-VAE, we set the dimension of both input and condition to 114 $\times$ 64, while the hidden and latent sizes are set to $1024$ and $2048$, respectively.
Our method is implemented by PyTorch and trained with the AdamW optimizer.
We train the model for 100 epochs on 4 NVIDIA 3090 GPUs with an initial
learning rate of $3e$-$5$ and a batch size of 40.

For comparison, 
We first select methods for individual tasks including Kalman Filter~(\textbf{KF})~\cite{kf}, \textbf{Seq2Seq}~\cite{seq2seq} and \textbf{Hotspots}~\cite{hotspots}. Additionally,
the methods for image/video understanding are chosen as the compared methods, which lack a joint mechanism.
The former includes \textbf{VIT}~\cite{dosovitskiy2020vit}, \textbf{HRNet}~\cite{hrnet}, \textbf{UNet}~\cite{unet}, \textbf{Segformer}~\cite{xie2021segformer}, while the latter contains \textbf{I3D}~\cite{I3D}, \textbf{Slowfast}~\cite{feichtenhofer2019slowfast}, \textbf{Timesformer}~\cite{timesformer}, \textbf{Uniformer-V2}~\cite{li2022uniformerv2}.
Furthermore, we select methods with high task relevance and joint design, including \textbf{FHOI}~\cite{liu2020forecasting} and \textbf{OCT}~\cite{liu2022joint}.
We conduct experiments using these approaches on Epic-kithens-100~\cite{Damen2018EPICKITCHENS}, EGO4D~\cite{grauman2022ego4d}, and EGTEA Gaze+~\cite{egtea} datasets.

\subsection{Evaluation Metrics}

\textbf{Hand Trajectory Evaluation.}
The predicted hand trajectories are referenced to the last observation frames.
\textbf{ADE}~\cite{liu2022joint} and \textbf{FDE}~\cite{liu2022joint} are chosen as hand trajectory evaluation metrics, which respectively represent the overall accuracy and endpoint accuracy of trajectories. Detailed descriptions are available in the supplementary materials.

\textbf{Interaction Hotspots Evaluation.}
For all datasets, we downsample each predicted heatmap to the size of 32$\times$32.
We choose \textbf{SIM}~\cite{SIM}, \textbf{AUC-J}~\cite{auc-j}, and \textbf{NSS}~\cite{nss} as interaction hotspots evaluation metrics. For a detailed explanation of the metrics, please refer to the supplementary materials.
\begin{table}[t]
\centering
\tabcolsep=0.055cm
\renewcommand\arraystretch{1.2}
\caption{\textbf{Results of different models.} Best results are depicted using \textcolor{red}{red} (sample 20 times) and \textcolor{dark_blue}{blue} (sample once), respectively. \textcolor{dark_green}{unilateral}, \textcolor{dark_green}{non-joint} and \textcolor{dark_green}{joint} denote prediction approaches for individual tasks, for two tasks without a collaborative design, and with a joint module, respectively. ``--'' denotes the vacancy of the value.  ``*'' represents the results obtained from 20 random samplings. $\uparrow$ and $\downarrow$ indicate a better performance when the value of the metric increases and decreases, respectively.}
\vspace{-8pt}
\setlength{\arrayrulewidth}{1.pt}
\resizebox{1.\linewidth}{!}{
\begin{tabular}{p{5mm}|c|ccccccccccccccccc}
\toprule

& \multicolumn{1}{c|}{\multirow{2}*{\textbf{Method}}}&  \multicolumn{5}{c|}{\textbf{EK100~\cite{Damen2018EPICKITCHENS}}}   & \multicolumn{5}{c|}{\textbf{EGO4D~\cite{grauman2022ego4d}}} & \multicolumn{5}{c}{\textbf{EGTEA~\cite{egtea}}} \\
&&\multicolumn{1}{c}{ADE $\downarrow$} &\multicolumn{1}{c}{FDE $\downarrow$} &\multicolumn{1}{c}{SIM $\uparrow$} &\multicolumn{1}{c}{AUC-J $\uparrow$} &\multicolumn{1}{c|}{NSS $\uparrow$}
&\multicolumn{1}{c}{ADE $\downarrow$} &\multicolumn{1}{c}{FDE $\downarrow$} &\multicolumn{1}{c}{SIM $\uparrow$} &\multicolumn{1}{c}{AUC-J $\uparrow$} &\multicolumn{1}{c|}{NSS $\uparrow$}
&\multicolumn{1}{c}{ADE $\downarrow$} &\multicolumn{1}{c}{FDE $\downarrow$} &\multicolumn{1}{c}{SIM $\uparrow$} &\multicolumn{1}{c}{AUC-J $\uparrow$} &\multicolumn{1}{c}{NSS $\uparrow$} \\
\midrule
\centering \multirow{4}{*}{\rotatebox{90}{\textbf{\textcolor{dark_green}{unilateral}}}}&\multicolumn{1}{r|}{{Center}} 
 
&\multicolumn{1}{c}{--}&\multicolumn{1}{c}{--} &\multicolumn{1}{c}{0.09} &\multicolumn{1}{c}{0.61} &\multicolumn{1}{c|}{0.33}
&\multicolumn{1}{c}{--} &\multicolumn{1}{c}{--} &\multicolumn{1}{c}{0.07} &\multicolumn{1}{c}{0.57} &\multicolumn{1}{c|}{0.15}
&\multicolumn{1}{c}{--} &\multicolumn{1}{c}{--} &\multicolumn{1}{c}{0.09} &\multicolumn{1}{c}{0.63} &\multicolumn{1}{c}{0.27}
\\

&\multicolumn{1}{r|}{{KF}~\cite{kf}}
&\multicolumn{1}{c}{0.33} &\multicolumn{1}{c}{0.32} &\multicolumn{1}{c}{--} &\multicolumn{1}{c}{--} &\multicolumn{1}{c|}{--}
&\multicolumn{1}{c}{--} &\multicolumn{1}{c}{--} &\multicolumn{1}{c}{--} &\multicolumn{1}{c}{--} &\multicolumn{1}{c|}{--}
&\multicolumn{1}{c}{0.49} &\multicolumn{1}{c}{0.48} &\multicolumn{1}{c}{--} &\multicolumn{1}{c}{--} &\multicolumn{1}{c}{--}
\\

&\multicolumn{1}{r|}{Seq2Seq~\cite{seq2seq}}
&\multicolumn{1}{c}{0.18} &\multicolumn{1}{c}{0.14} &\multicolumn{1}{c}{--} &\multicolumn{1}{c}{--} &\multicolumn{1}{c|}{--}
&\multicolumn{1}{c}{--} &\multicolumn{1}{c}{--} &\multicolumn{1}{c}{--} &\multicolumn{1}{c}{--} &\multicolumn{1}{c|}{--}
&\multicolumn{1}{c}{0.18} &\multicolumn{1}{c}{0.14} &\multicolumn{1}{c}{--} &\multicolumn{1}{c}{--} &\multicolumn{1}{c}{--}
\\
&\multicolumn{1}{r|}{Hotspots~\cite{hotspots}} 
&\multicolumn{1}{c}{--} &\multicolumn{1}{c}{--} &\multicolumn{1}{c}{0.15} &\multicolumn{1}{c}{0.66} &\multicolumn{1}{c|}{0.53}
&\multicolumn{1}{c}{--} &\multicolumn{1}{c}{--} &\multicolumn{1}{c}{--} &\multicolumn{1}{c}{--} &\multicolumn{1}{c|}{--}
&\multicolumn{1}{c}{--} &\multicolumn{1}{c}{--} &\multicolumn{1}{c}{0.15} &\multicolumn{1}{c}{0.71} &\multicolumn{1}{c}{0.69}
\\
\midrule

\centering \multirow{9}{*}{\rotatebox{90}{ \textbf{\textcolor{dark_green}{non-joint}}}}&\multicolumn{1}{r|}{{I3D}~\cite{I3D}}  
&\multicolumn{1}{c}{0.22} &\multicolumn{1}{c}{0.23} &\multicolumn{1}{c}{0.13} &\multicolumn{1}{c}{0.56} &\multicolumn{1}{c|}{0.46}
&\multicolumn{1}{c}{0.22} &\multicolumn{1}{c}{0.22} &\multicolumn{1}{c}{0.12} &\multicolumn{1}{c}{0.57} &\multicolumn{1}{c|}{0.42}
&\multicolumn{1}{c}{0.23} &\multicolumn{1}{c}{0.22} &\multicolumn{1}{c}{0.13} &\multicolumn{1}{c}{0.60} &\multicolumn{1}{c}{0.49}
\\ 
 
&\multicolumn{1}{r|}{{SLOWFAST}~\cite{feichtenhofer2019slowfast}} 
&\multicolumn{1}{c}{0.22} &\multicolumn{1}{c}{0.24} &\multicolumn{1}{c}{0.14} &\multicolumn{1}{c}{0.62} &\multicolumn{1}{c|}{0.47}
&\multicolumn{1}{c}{0.23} &\multicolumn{1}{c}{0.22} &\multicolumn{1}{c}{0.13} &\multicolumn{1}{c}{0.58} &\multicolumn{1}{c|}{0.44}
&\multicolumn{1}{c}{0.21} &\multicolumn{1}{c}{0.23} &\multicolumn{1}{c}{0.14} &\multicolumn{1}{c}{0.58} &\multicolumn{1}{c}{0.46}
\\

&\multicolumn{1}{r|}{{Timesformer}~\cite{timesformer}}
&\multicolumn{1}{c}{0.23} &\multicolumn{1}{c}{0.23} &\multicolumn{1}{c}{0.14} &\multicolumn{1}{c}{0.60} &\multicolumn{1}{c|}{0.42}
&\multicolumn{1}{c}{0.24} &\multicolumn{1}{c}{0.24} &\multicolumn{1}{c}{0.12} &\multicolumn{1}{c}{0.59} &\multicolumn{1}{c|}{0.40}
&\multicolumn{1}{c}{0.21} &\multicolumn{1}{c}{0.22} &\multicolumn{1}{c}{0.14} &\multicolumn{1}{c}{0.58} &\multicolumn{1}{c}{0.45}
\\

&\multicolumn{1}{r|}{{Uniformer-V2}~\cite{li2022uniformerv2}}
&\multicolumn{1}{c}{0.20} &\multicolumn{1}{c}{0.20} &\multicolumn{1}{c}{0.13} &\multicolumn{1}{c}{0.59} &\multicolumn{1}{c|}{0.43}
&\multicolumn{1}{c}{0.21} &\multicolumn{1}{c}{0.22} &\multicolumn{1}{c}{0.12} &\multicolumn{1}{c}{0.62} &\multicolumn{1}{c|}{0.40}
&\multicolumn{1}{c}{0.20} &\multicolumn{1}{c}{0.21} &\multicolumn{1}{c}{0.13} &\multicolumn{1}{c}{0.58} &\multicolumn{1}{c}{0.46}
\\

&\multicolumn{1}{r|}{{UNet}~\cite{unet}} 
&\multicolumn{1}{c}{0.19} &\multicolumn{1}{c}{0.18} &\multicolumn{1}{c}{0.15} &\multicolumn{1}{c}{0.62} &\multicolumn{1}{c|}{0.57}
&\multicolumn{1}{c}{0.18} &\multicolumn{1}{c}{0.18} &\multicolumn{1}{c}{0.14} &\multicolumn{1}{c}{0.59} &\multicolumn{1}{c|}{0.55}
&\multicolumn{1}{c}{0.18} &\multicolumn{1}{c}{0.18} &\multicolumn{1}{c}{0.14} &\multicolumn{1}{c}{0.60} &\multicolumn{1}{c}{0.60}
\\

&\multicolumn{1}{r|}{{HRNet}~\cite{hrnet}}
&\multicolumn{1}{c}{0.18} &\multicolumn{1}{c}{0.19} &\multicolumn{1}{c}{0.16} &\multicolumn{1}{c}{0.62} &\multicolumn{1}{c|}{0.63}
&\multicolumn{1}{c}{0.19} &\multicolumn{1}{c}{0.20} &\multicolumn{1}{c}{0.14} &\multicolumn{1}{c}{0.62} &\multicolumn{1}{c|}{0.58}
&\multicolumn{1}{c}{0.18} &\multicolumn{1}{c}{0.18} &\multicolumn{1}{c}{0.16} &\multicolumn{1}{c}{0.63} &\multicolumn{1}{c}{0.65}
\\
&\multicolumn{1}{r|}{{VIT}~\cite{dosovitskiy2020vit}}
&\multicolumn{1}{c}{0.18} &\multicolumn{1}{c}{0.19} &\multicolumn{1}{c}{0.14} &\multicolumn{1}{c}{0.65} &\multicolumn{1}{c|}{0.63}
&\multicolumn{1}{c}{0.21} &\multicolumn{1}{c}{0.21} &\multicolumn{1}{c}{0.13} &\multicolumn{1}{c}{0.59} &\multicolumn{1}{c|}{0.45}
&\multicolumn{1}{c}{0.19} &\multicolumn{1}{c}{0.19} &\multicolumn{1}{c}{0.15} &\multicolumn{1}{c}{0.63} &\multicolumn{1}{c}{0.61}
\\

&\multicolumn{1}{r|}{{CrossVIT}~\cite{chen2021crossvit}} 
&\multicolumn{1}{c}{0.19} &\multicolumn{1}{c}{0.20} &\multicolumn{1}{c}{0.14} &\multicolumn{1}{c}{0.62} &\multicolumn{1}{c|}{0.59}
&\multicolumn{1}{c}{0.19} &\multicolumn{1}{c}{0.20} &\multicolumn{1}{c}{0.13} &\multicolumn{1}{c}{0.60} &\multicolumn{1}{c|}{0.55}
&\multicolumn{1}{c}{0.20} &\multicolumn{1}{c}{0.21} &\multicolumn{1}{c}{0.13} &\multicolumn{1}{c}{0.59} &\multicolumn{1}{c}{0.49}
\\

&\multicolumn{1}{r|}{{Segformer-b2}~\cite{xie2021segformer}}
&\multicolumn{1}{c}{0.17} &\multicolumn{1}{c}{0.17} &\multicolumn{1}{c}{0.16} &\multicolumn{1}{c}{0.64} &\multicolumn{1}{c|}{0.66}
&\multicolumn{1}{c}{0.16} &\multicolumn{1}{c}{0.16} &\multicolumn{1}{c}{0.15} &\multicolumn{1}{c}{0.64} &\multicolumn{1}{c|}{0.62}
&\multicolumn{1}{c}{0.17} &\multicolumn{1}{c}{0.17} &\multicolumn{1}{c}{0.16} &\multicolumn{1}{c}{0.63} &\multicolumn{1}{c}{0.65}
\\
\midrule

\centering \multirow{5}{*}{\rotatebox{90}{ \textbf{\textcolor{dark_green}{joint}}}}&\multicolumn{1}{r|}{FHOI~\cite{liu2020forecasting}} 
&\multicolumn{1}{c}{0.36} &\multicolumn{1}{c}{0.35} &\multicolumn{1}{c}{0.10} &\multicolumn{1}{c}{0.55} &\multicolumn{1}{c|}{0.39}
&\multicolumn{1}{c}{--} &\multicolumn{1}{c}{--} &\multicolumn{1}{c}{--} &\multicolumn{1}{c}{--} &\multicolumn{1}{c|}{--}
&\multicolumn{1}{c}{0.34} &\multicolumn{1}{c}{0.34} &\multicolumn{1}{c}{0.12} &\multicolumn{1}{c}{0.63} &\multicolumn{1}{c}{0.42}
\\

&\multicolumn{1}{r|}{OCT~\cite{liu2022joint}} 
&\multicolumn{1}{c}{0.15} &\multicolumn{1}{c}{0.17} &\multicolumn{1}{c}{--} &\multicolumn{1}{c}{--} &\multicolumn{1}{c|}{--}
&\multicolumn{1}{c}{0.15} &\multicolumn{1}{c}{0.16} &\multicolumn{1}{c}{--} &\multicolumn{1}{c}{--} &\multicolumn{1}{c|}{--}
&\multicolumn{1}{c}{0.16} &\multicolumn{1}{c}{0.16} &\multicolumn{1}{c}{--} &\multicolumn{1}{c}{--} &\multicolumn{1}{c}{--}

\\
&\multicolumn{1}{r|}{OCT*~\cite{liu2022joint}} 
&\multicolumn{1}{c}{0.12} &\multicolumn{1}{c}{0.11} &\multicolumn{1}{c}{0.19} &\multicolumn{1}{c}{0.72} &\multicolumn{1}{c|}{0.72}
&\multicolumn{1}{c}{0.12} &\multicolumn{1}{c}{0.12} &\multicolumn{1}{c}{0.22} &\multicolumn{1}{c}{0.73} &\multicolumn{1}{c|}{0.88}
&\multicolumn{1}{c}{0.14} &\multicolumn{1}{c}{0.14} &\multicolumn{1}{c}{0.23} &\multicolumn{1}{c}{0.75} &\multicolumn{1}{c}{1.01}

\\
&\multicolumn{1}{r|}{{Ours}} 
&\multicolumn{1}{c}{{\textcolor{dark_blue}{\textbf{0.12}}}} &\multicolumn{1}{c}{{\textcolor{dark_blue}{\textbf{0.14}}}} &\multicolumn{1}{c}{{--}} &\multicolumn{1}{c}{{--}} &\multicolumn{1}{c|}{{--}}
&\multicolumn{1}{c}{{\textcolor{dark_blue}{\textbf{0.11}}}} &\multicolumn{1}{c}{{\textcolor{dark_blue}{\textbf{0.13}}}} &\multicolumn{1}{c}{{--}} &\multicolumn{1}{c}{{--}} &\multicolumn{1}{c|}{{--}}
&\multicolumn{1}{c}{{\textcolor{dark_blue}{\textbf{0.12}}}} &\multicolumn{1}{c}{{\textcolor{dark_blue}{\textbf{0.14}}}} &\multicolumn{1}{c}{{--}} &\multicolumn{1}{c}{{--}} &\multicolumn{1}{c}{{--}}
\\

&\multicolumn{1}{r|}{{Ours*}} 
&\multicolumn{1}{c}{\textcolor{red}{\textbf{0.10}}} &\multicolumn{1}{c}{\textcolor{red}{\textbf{0.08}}} &\multicolumn{1}{c}{\textcolor{red}{\textbf{0.34}}} &\multicolumn{1}{c}{\textcolor{red}{\textbf{0.81}}}&\multicolumn{1}{c|}{\textcolor{red}{\textbf{2.07}}}
&\multicolumn{1}{c}{\textcolor{red}{\textbf{0.09}}} &\multicolumn{1}{c}{\textcolor{red}{\textbf{0.08}}} &\multicolumn{1}{c}{\textcolor{red}{\textbf{0.32}}} &\multicolumn{1}{c}{\textcolor{red}{\textbf{0.77}}} &\multicolumn{1}{c|}{\textcolor{red}{\textbf{1.70}}}
&\multicolumn{1}{c}{\textcolor{red}{\textbf{0.09}}} &\multicolumn{1}{c}{\textcolor{red}{\textbf{0.08}}} &\multicolumn{1}{c}{\textcolor{red}{\textbf{0.32}}} &\multicolumn{1}{c}{\textcolor{red}{\textbf{0.80}}} &\multicolumn{1}{c}{\textcolor{red}{\textbf{2.03}}}
\\

\bottomrule
\end{tabular}
}
\label{tab:t1}
\vspace{-3mm}
\end{table}

\subsection{Quantitative and Qualitative Comparisons}
\label{sec:4.4}


As shown in Tab.~\ref{tab:t1}, our method achieves state-of-the-art performance in both trajectory prediction and interaction hotspots anticipation, with comparable leading margins across all three datasets. Taking the EGO4D dataset~\cite{grauman2022ego4d} with the richest variety of scenes as an example. In trajectory prediction, to ensure fair comparison, we additionally present results for OCT and BOT with only one sampling iteration. 
Under this single-sampling condition, BOT outperforms Segformer-b2, the best method in image/video understanding, with respective leads of 31.3\% and 18.8\% in ADE and FDE. Compared to OCT, BOT maintains advantages of 26.7\% and 18.8\% in ADE and FDE, respectively. These advantages persist even with 20 samplings, while the lead in FDE expands to 33.3\%, underscoring the rationale behind BOT's stochastic sampling mechanism. Regarding interaction hotspots, we achieve the best results across all metrics. Specifically, BOT surpasses the second-best approach (OCT) by 45.5\% in SIM, 5.5\% in AUC-J, and 93.2\% in NSS, respectively. In summary, our method leverages the intrinsic correlations between hand trajectories and interaction hotspots more effectively, enhancing overall prediction accuracy of interaction intention, and exhibiting robustness across the three benchmark datasets.
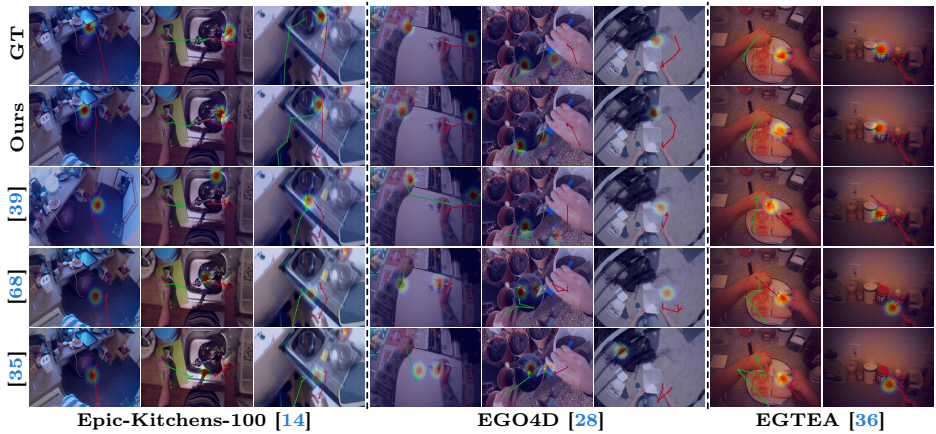
\begin{figure}[t]
        \centering
            \begin{overpic}[width=1.\linewidth]{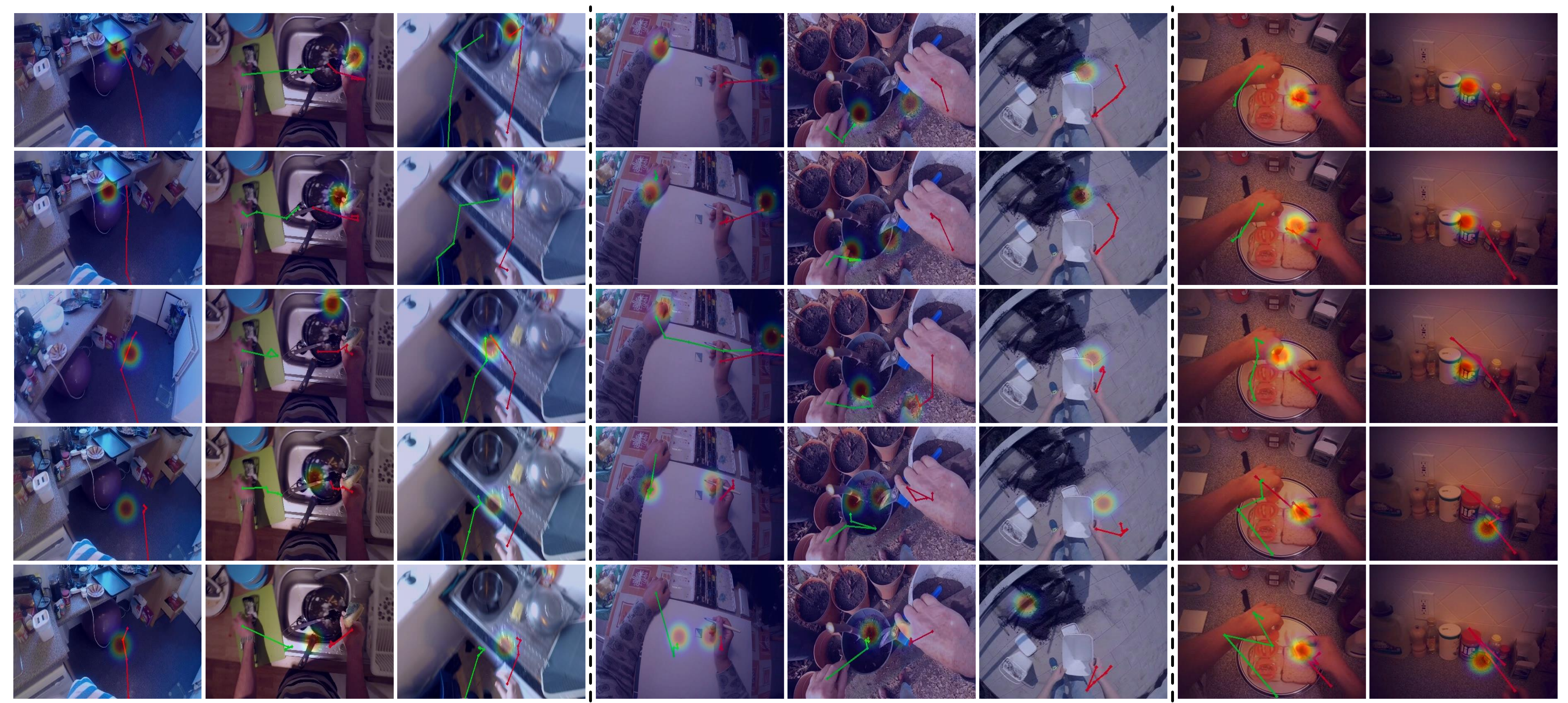}
    \put(-1.2,38){\rotatebox{90}{\scriptsize\textbf{GT}}}
    \put(-1.2,28){\rotatebox{90}{\scriptsize\textbf{Ours}}}
    \put(-1.5,19.5){\rotatebox{90}{\scriptsize\textbf{\cite{liu2022joint}}}}
    \put(-1.5,10.5){\rotatebox{90}{\scriptsize\textbf{\cite{xie2021segformer}}}}
    \put(-1.5,1.5){\rotatebox{90}
    {\scriptsize\textbf{\cite{li2022uniformerv2}}}}
    \put(6,-1.8){\rotatebox{0}{\scriptsize\textbf{Epic-Kitchens-100~\cite{Damen2018EPICKITCHENS}}}}
    \put(49.65,-1.8){\rotatebox{0}{\scriptsize\textbf{EGO4D~\cite{grauman2022ego4d}}}}
    \put(80,-1.8){\rotatebox{0}{\scriptsize\textbf{EGTEA~\cite{egtea}}}}
        \end{overpic}
    \caption{\textbf{Visualization of future hand trajectories and interaction hotspots on EK100, EGO4D and EGETA}.
    We show the visualization results of our model, the joint prediction model (OCT)~\cite{liu2022joint}, the segmentation model (Segformer-b2)~\cite{xie2021segformer}, and the video understanding model (Uniformer-V2)~\cite{li2022uniformerv2}.
    The \textcolor{green}{green} represents the trajectory of the left hand, while the \textcolor{red}{red} denotes the trajectory of the right hand.
    } 
    \label{fig:keshihua}
    \vspace{-6mm}
\end{figure}

\begin{wraptable}{r}{0.38\textwidth}
\centering
\scriptsize
    \vspace{-10mm}
    \caption{\textbf{Inference time and parameter count}. ``*'' represents inference time obtained from 20 random samplings.}
    \begin{tabular}{c|cc}
\toprule
\multicolumn{1}{c|}{\textbf{Method}}& \multicolumn{1}{c}{\textbf{Inf. time}}& \multicolumn{1}{c}{\textbf{Param}}\\
\midrule
\multicolumn{1}{l|}{Segformer-b2~\cite{xie2021segformer}}& \multicolumn{1}{c}{21ms}& \multicolumn{1}{c}{2.7M} \\
\multicolumn{1}{l|}{Uniformer-V2~\cite{li2022uniformerv2}}& \multicolumn{1}{c}{310ms}& \multicolumn{1}{c}{35.4M} \\
\multicolumn{1}{l|}{OCT~\cite{liu2022joint}}& \multicolumn{1}{c}{52ms}& \multicolumn{1}{c}{3.9M} \\
\multicolumn{1}{l|}{OCT*~\cite{liu2022joint}}& \multicolumn{1}{c}{923ms}& \multicolumn{1}{c}{3.9M} \\
\rowcolor{gray!20}
\multicolumn{1}{l|}{Ours}& \multicolumn{1}{c}{55ms}& \multicolumn{1}{c}{14.3M} \\
\rowcolor{gray!20}
\multicolumn{1}{l|}{Ours*}& \multicolumn{1}{c}{92ms}& \multicolumn{1}{c}{14.3M} \\
    \bottomrule
    \end{tabular}
    \label{tab:time}
    \vspace{-5.5mm}
\end{wraptable} 
To investigate the efficiency of the model, we provide statistics on inference time and parameter count (Tab.~\ref{tab:time}). For each method, we test the inference time of one sample on a single NVIDIA 3090 GPU. In comparison to OCT, although our model has a larger parameter count, it exhibits a similar inference time when sampling once. With multiple samplings, our model displays a notable advantage in inference time. This can be attributed to the improved efficiency of the Stochastic Unit in BOT, obviating the necessity for repetitive inference of complete trajectories compared to OCT. Given BOT's superior inference capability, our approach demonstrates enhanced efficiency.

To further explore the subjective outcomes of models, we report the visualization results of three datasets, as shown in Fig.~\ref{fig:keshihua}. Compared to other methods, our model provides more accurate prediction results in both hand trajectories and interaction hotspots. When there is significant movement of hands (\textbf{column 1,3,8}), the BOT's predictions exhibit the highest accuracy. It indicates that our Spatial-Temporal Reconstruction Module can efficiently utilize the human motion information to exploit the hand interaction intention and thus accurately predict the hand trajectory. In cases where there are no identifiable interactive objects in the field of view (\textbf{column 6}), our method anticipates more accurate interaction intentions. It illustrates that our Bi-Progressive mechanism can eliminate the uncertainty in the interaction process by mining the intrinsic connection between the hand trajectory and the contact points, which results in a more accurate inference in the interaction region and the hand trajectory. Moreover, our predictions closely align with the actual scenario when both hands simultaneously exhibit trajectories and interaction hotspots (\textbf{columns 4,5}), which suggests the strategy of differentiating between left and right hands during the prediction of interaction intentions, thereby mitigating cross-hand interference. More visualization results are available in the supplementary materials.

\subsection{Ablation Study}

\begin{figure}[t]
    \begin{minipage}{0.48\linewidth}
    \centering
     \includegraphics[width=0.99\linewidth]{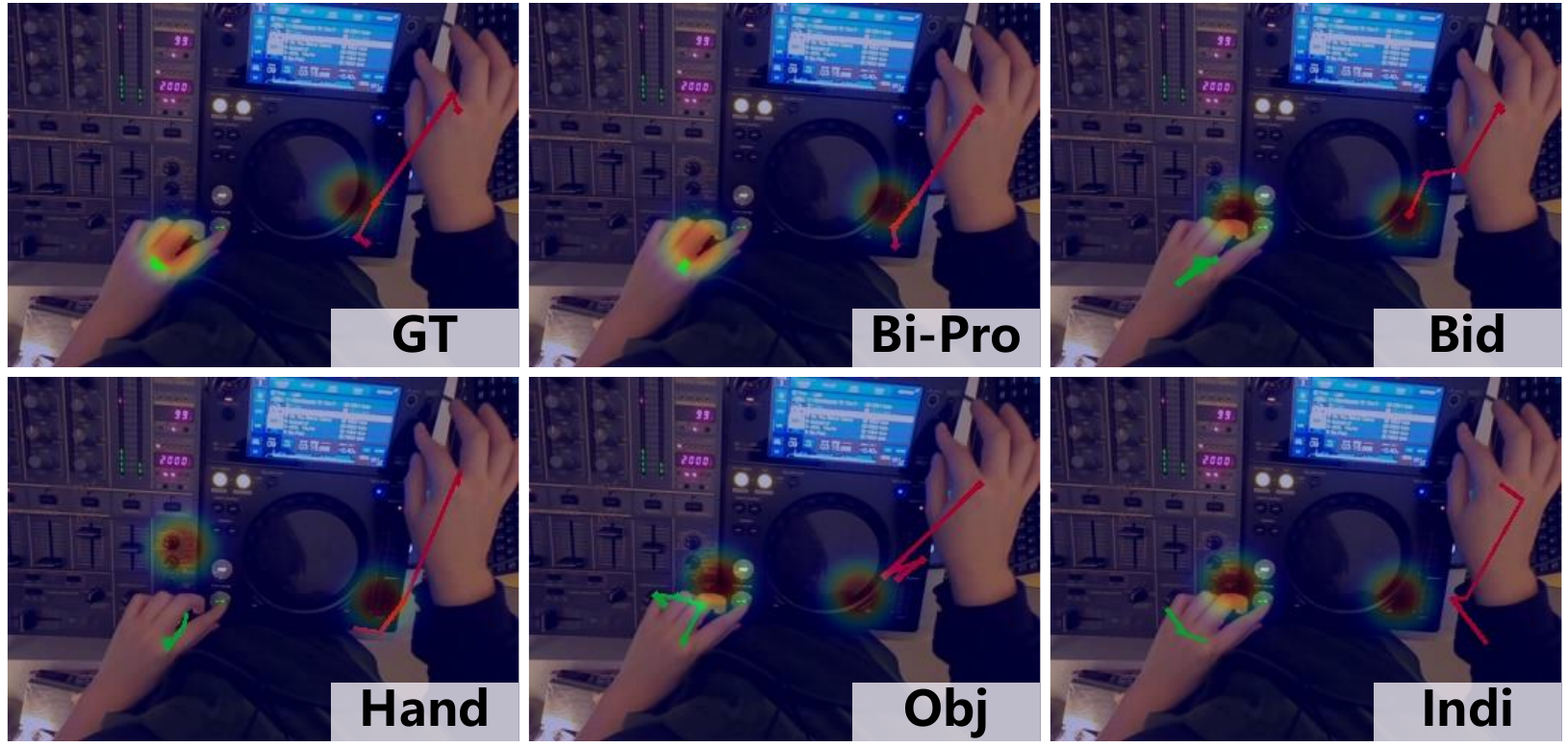}
    \caption{\textbf{Visualization results of different joint approaches.} Based on the order from left to right and top to bottom, the corresponding methods are \textbf{GT}, \textbf{Bi-Progressive}, \textbf{Bidirectional}, \textbf{Hand}, \textbf{Object}, \textbf{Individual}.}
    \label{fig:ablation1}
    \end{minipage}
    \hfill
    \begin{minipage}{0.48\linewidth}
    \vspace{-1pt}
    \centering
     \includegraphics[width=0.99\linewidth]{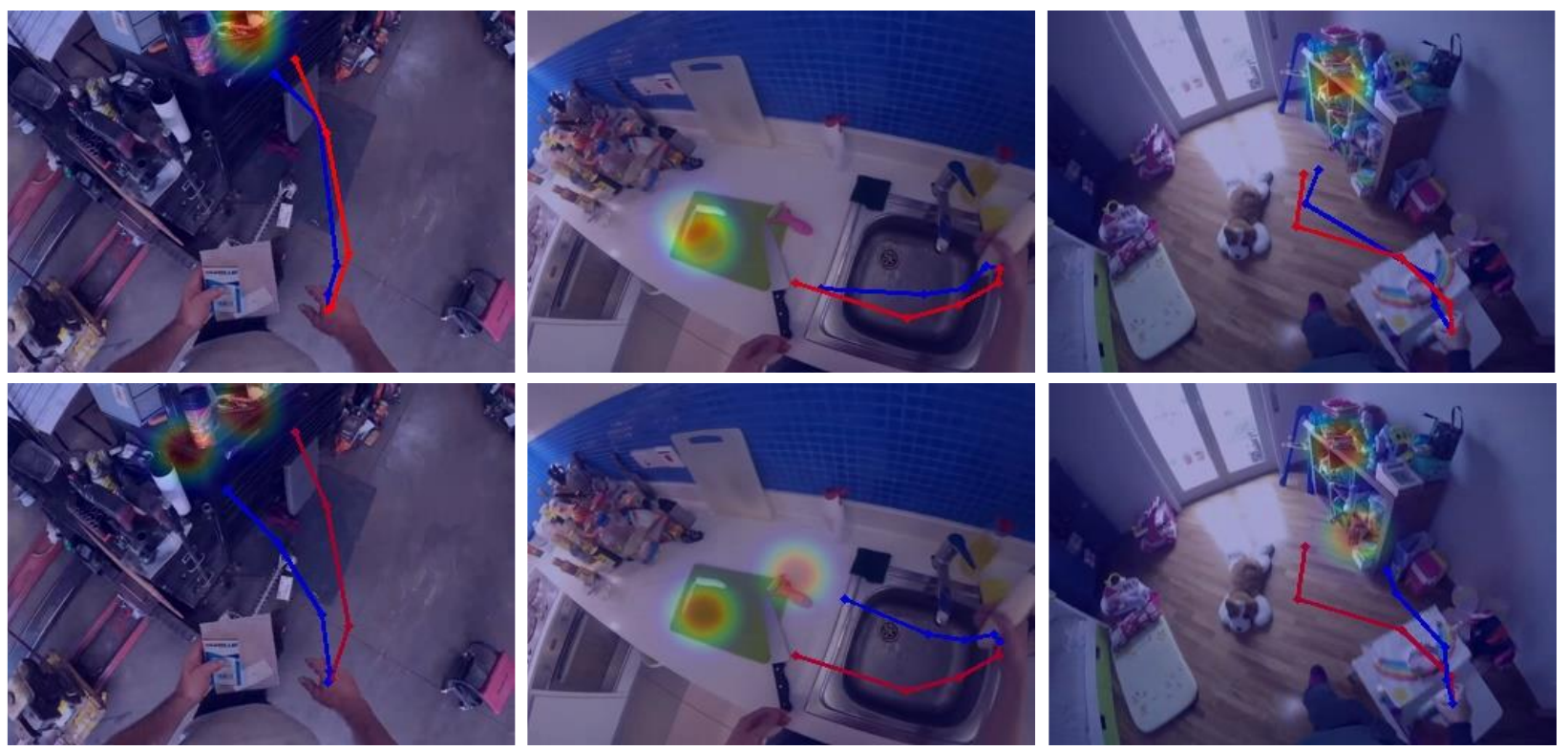}
     \put(0.2,71.5){\rotatebox{270}{\scriptsize\textbf{Bi-Pro}}}
    \put(0.7,26){\rotatebox{270}{\scriptsize\textbf{Bid}}}
    \caption{\textbf{The role of Progressive Mechanism}. The \textcolor{red}{red} refers to the ground truth, while the \textcolor{blue}{blue} represents the predictions. From top to bottom, \textbf{Bi-Pro} stands for Bi-Progressive while \textbf{Bid} refers to bidirectional.}
    \label{fig:sub}
    \end{minipage}
\end{figure}

\textbf{Different joint approaches.} To explore the impact of the Bi-Progressive mechanism on model performance, we experiment with five different joint configurations: \textbf{Individual}, \textbf{Hand}, \textbf{Object}, \textbf{Bidirectional} and \textbf{Bi-Progressive}.
The results are shown in Tab.~\ref{tab:ablation1} and Fig.~\ref{fig:ablation1}. It suggests that the network is challenging to accurately predict trajectories and interaction hotspots during individual predictions, and the separate incorporation of hands and objects can effectively improve the prediction accuracy of interaction hotspots. With the combination of hand and object information, the prediction performance of the trajectory and interaction region is greatly improved. However, there is still a gap compared to the Bi-Progressive mechanism. Further, Fig.~\ref{fig:sub} shows that the introduction of the progressive mechanism plays a pivotal role in mitigating the 
 accumulation of prediction errors over time.

\begin{figure}[]
\centering
\scriptsize
    \vspace{-1mm}
    \centering
    \includegraphics[width=0.9\linewidth]{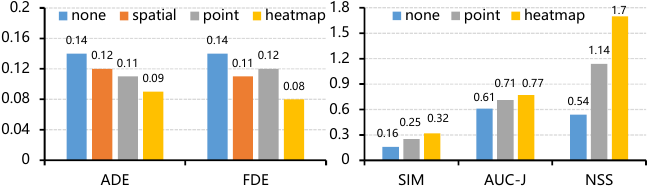}
    \vspace{4mm}
    \caption{\textbf{Different conditions in dual prediction branches.} The left displays conditions in the trajectory prediction branch. The right shows conditions in the interaction hotspots prediction branch.}
    \label{fig:ablation2}
    
\end{figure}

\textbf{Conditions in dual branches.} The left portion in Fig.~\ref{fig:ablation2} delineates various conditions for the trajectory prediction branch. \textbf{None} signifies the absence of the fusion between two consecutive trajectory points, \textbf{spatial} denotes the exchange between the main and the condition input, \textbf{point} and \textbf{heatmap} represent different \begin{wraptable}{r}{0.57\textwidth}
    \centering
    \scriptsize
    \vspace{-9mm}
    \caption{\textbf{Results of different joint forecasting methods.} \textbf{Individual} refers to separately predict two subtasks. \textbf{Hand} indicates that the hand trajectory is the condition of the interaction hotspots. \textbf{Object} stands for correcting the trajectory using predicted contact points.
    \textbf{Bidirectional} combines the methods \textbf{Hand} and \textbf{Object}.}
    
    \begin{tabular}{c|ccccc}
    \toprule
       \multicolumn{1}{c|}{\multirow{2}*{\textbf{Method}}}  & \multicolumn{2}{c|}{\textbf{Trajectory}}
      & \multicolumn{3}{c}{\textbf{Interaction Hotspots}} \\
      \cmidrule(lr){2-3}\cmidrule(lr){4-6}
      &\multicolumn{1}{c}{ADE $\downarrow$} &\multicolumn{1}{c|}{FDE $\downarrow$} &\multicolumn{1}{c}{SIM $\uparrow$} &\multicolumn{1}{c}{AUC-J $\uparrow$} &\multicolumn{1}{c}{NSS $\uparrow$} \\
      \midrule
      \multicolumn{1}{c|}{Individual} & \multicolumn{1}{c}{0.14}& \multicolumn{1}{c|}{0.14}
      & \multicolumn{1}{c}{0.18} & \multicolumn{1}{c}{0.68} & \multicolumn{1}{c}{0.86} \\
      \multicolumn{1}{c|}{Hand} & \multicolumn{1}{c}{0.13}& \multicolumn{1}{c|}{0.12}
      & \multicolumn{1}{c}{0.25} & \multicolumn{1}{c}{0.72} & \multicolumn{1}{c}{1.19} \\
      \multicolumn{1}{c|}{Object} & \multicolumn{1}{c}{0.12}& \multicolumn{1}{c|}{0.12}
      & \multicolumn{1}{c}{0.23} & \multicolumn{1}{c}{0.70} & \multicolumn{1}{c}{1.04} \\
      \multicolumn{1}{c|}{Bidirectional} & \multicolumn{1}{c}{0.10}& \multicolumn{1}{c|}{0.09}
      & \multicolumn{1}{c}{0.28} & \multicolumn{1}{c}{0.73} & \multicolumn{1}{c}{1.48} \\
      \rowcolor{gray!20}
       \multicolumn{1}{c|}{\textbf{Bi-Progressive}} &\multicolumn{1}{c}{\textbf{0.09}} &\multicolumn{1}{c|} {\textbf{0.08}} &\multicolumn{1}{c}{\textbf{0.32}} &\multicolumn{1}{c}{\textbf{0.77}} &\multicolumn{1}{c}{\textbf{1.70}}\\
       \bottomrule
    \end{tabular}
    
    \label{tab:ablation1}
    \vspace{-5mm}
\end{wraptable}modes of the condition, 
where the former is a coordinate and the latter is a heatmap, incorporating the potential distribution of the hand position. Experimental results indicate that the cross-attention block contributes to the temporal continuity of the trajectory, and global spatial features exert a more substantial influence on the prediction. Furthermore, employing an appropriate region to
 depict the hand position contributes more significantly to the propagation of trajectory features.
The right side of Fig.~\ref{fig:ablation2} illustrates conditions within the interaction hotspots prediction branch. It is evident that using \textbf{heatmap} as the condition for the C-VAE yields superior results compared to \textbf{point}. This implies that the contact point also exhibits a degree of uncertainty within a specific range.

\begin{figure}[]
\centering
\footnotesize
    \vspace{-5mm}
    \centering
    \includegraphics[width=0.9\linewidth]{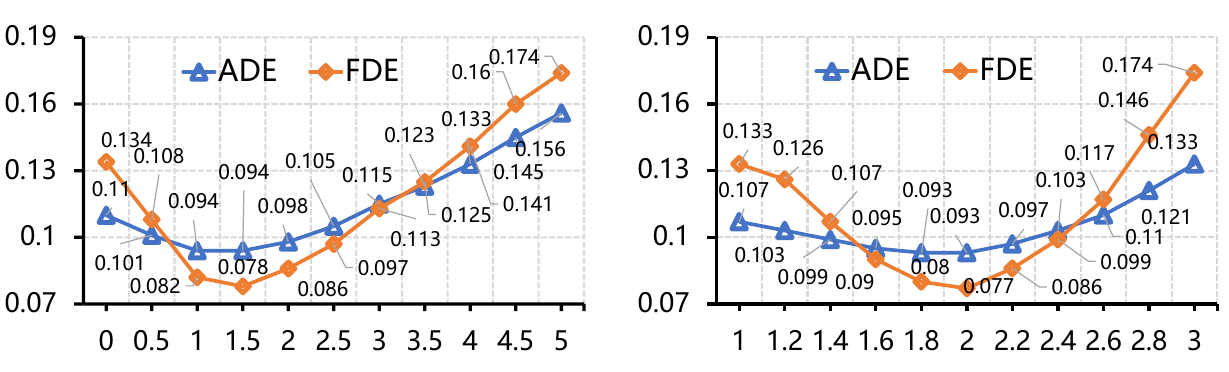}
    \vspace{4mm}
    \put(-230,-4.6){$\alpha $}
    \put(-70,-5.5){$\beta $}
    \vspace{-3mm}
    \caption{\textbf{Uncertainty in hand trajectory forecasting}. As  $\alpha$ and $\beta$ increase, they denote a heightened degree of overall trajectory uncertainty.}
    \label{fig:ablation3}
    
\end{figure}
\textbf{Uncertainty in anticipation.} Uncertainty is a crucial consideration in interaction intention anticipation. Following the logic of natural human behavior~\cite{uncetainty1,uncertainty2,uncertainty3}, both hand trajectories~\cite{mangalam2020not,gupta2018social,uncertain10} and interaction hotspots~\cite{liu2022joint, Luo_2022_CVPR,luo2023learning} predictions inherently possess a degree of uncertainty. 
\begin{figure}[t]
    \begin{subfigure}{0.49\linewidth}
    \centering
    \includegraphics[width=0.99\linewidth]{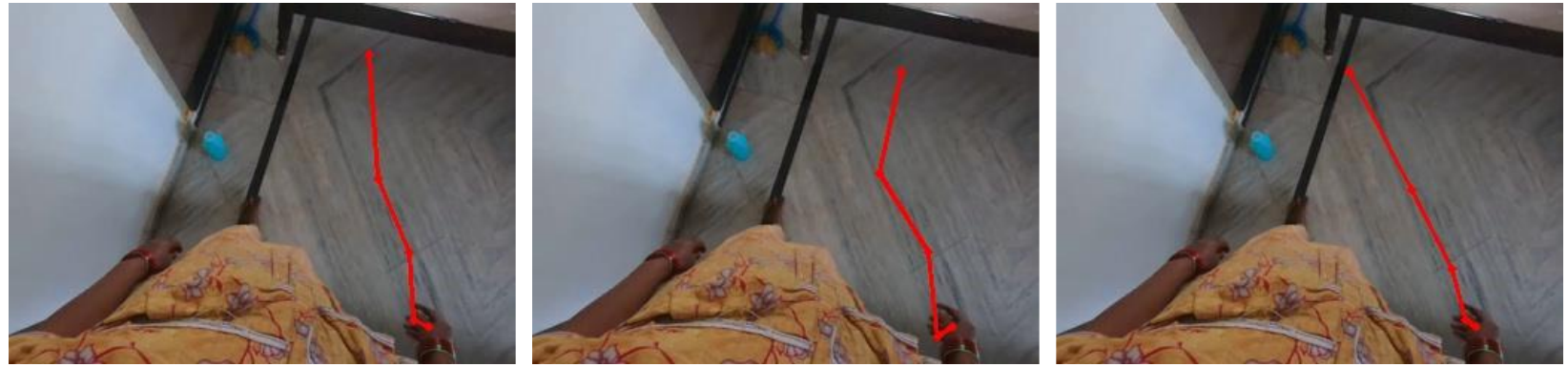}
    \caption{\textbf{Long-length trajectories}}
    \label{fig:last1}
    \end{subfigure}
    \hfill
    \begin{subfigure}{0.49\linewidth}
    \centering
    \includegraphics[width=0.99\linewidth]{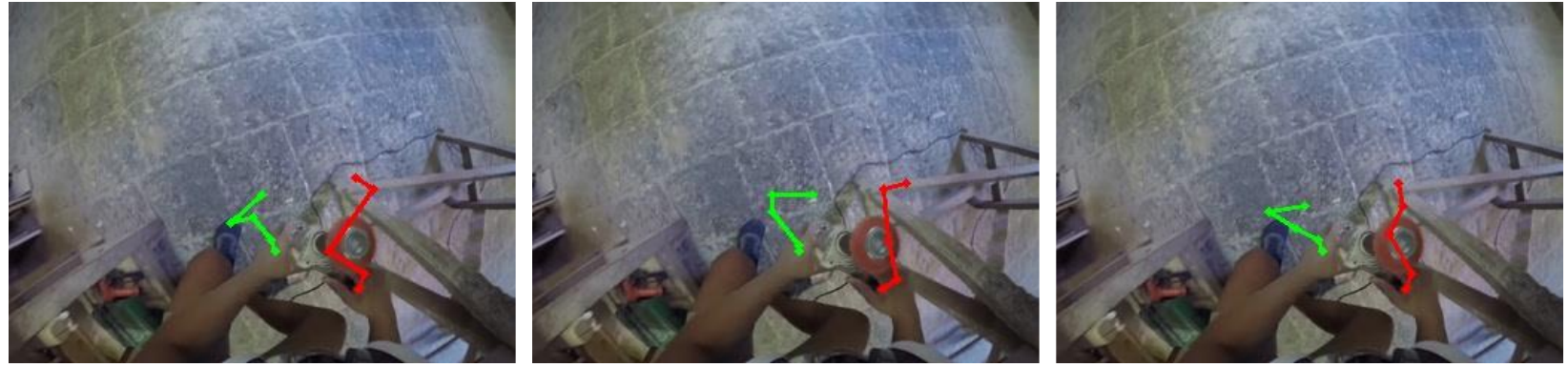}
    \caption{\textbf{Medium-length trajectories}}
    \label{fig:last2}
    \end{subfigure}
    
    \begin{subfigure}{0.49\linewidth}
    \centering
    \includegraphics[width=0.99\linewidth]{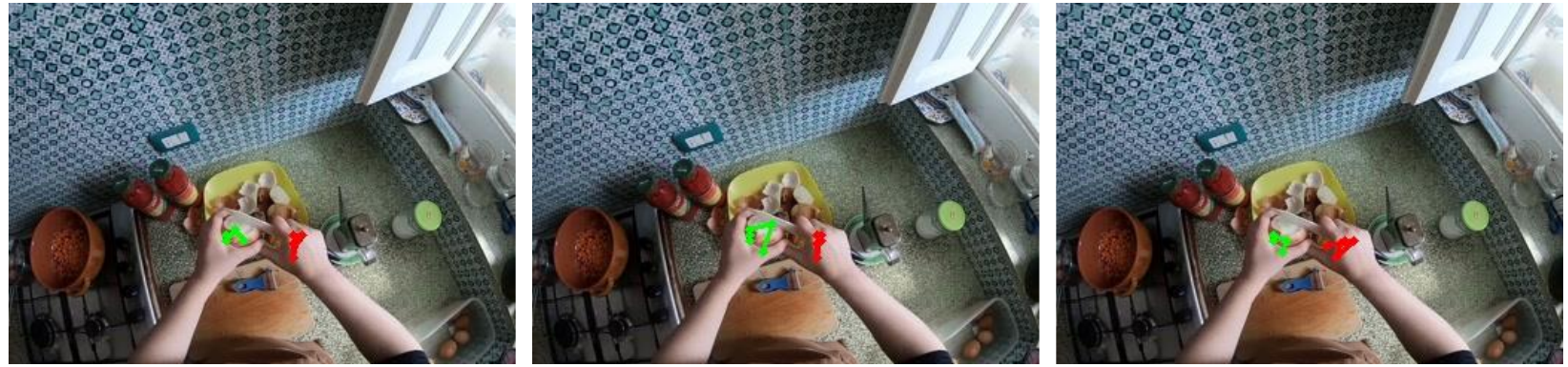}
    \caption{\textbf{Short-length trajectories}}
    \label{fig:last3}
    \end{subfigure}
    \hfill
    \begin{subfigure}{0.49\linewidth}
    \centering
    \includegraphics[width=0.99\linewidth]{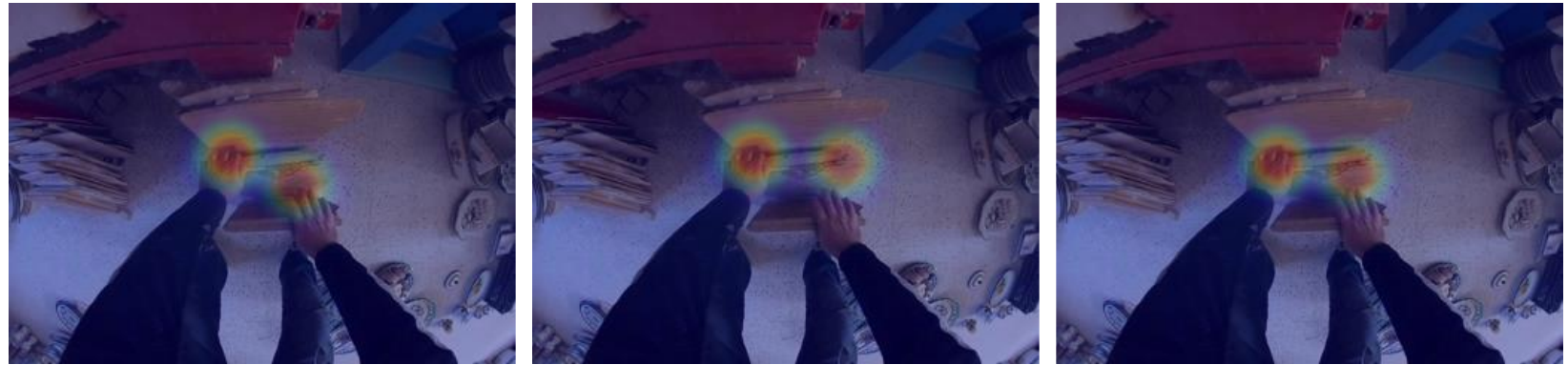}
    \caption{\textbf{interaction hotspots}}
    \label{fig:last4}
    \end{subfigure}
    
    \caption{\textbf{Uncertainty in anticipation}. (a), (b) and (c) respectively depict long-length trajectories, medium-length trajectories, and short-length trajectories, while (d) illustrates the uncertainty of interaction hotspots.}
    \label{fig:last}
    \vspace{-2mm}
\end{figure}
To investigate the impact of uncertain region scale and the magnitude of uncertainty variations in hand trajectories, we modify $\alpha$ and $\beta$ in Eq.~\ref{eq:16}, with the resultant outcomes depicted in Fig.~\ref{fig:ablation3}.
$\alpha$ denotes the overall scale of the uncertain region, when $\alpha$ equals 0, the region's size is zero, and the network output yields a unique result.
As $\alpha$ increases, the uncertainty of trajectories grows, leading to improved accuracy due to random samplings. However, given the limitations in sampling times, as $\alpha$ continues to increase, the accuracy naturally decreases. 
$\beta$ determines the magnitude of trajectory uncertainty increasing from front to back, when $\beta$ equals 1, each trajectory point has the same level of uncertainty. Considering the escalating complexity and uncertainty in trajectory prediction over time, an elevation in $\beta$ initially corresponds to augmented accuracy, followed by a subsequent decline.
This gradual increase in uncertainty also renders the FDE more sensitive to $\alpha$ and $\beta$, as shown in Fig.~\ref{fig:last1}-\ref{fig:last3}.
Regarding interaction hotspots, \textbf{BOT} exhibits relatively small uncertainty (Fig.~\ref{fig:last4}), as for a specific intention, the interaction area becomes increasingly influenced by object constraints. This is achieved by tightly controlling the uncertainty within a reasonably small range through the process of enabling C-VAE to generate feature maps and subsequently select points, thereby ensuring the network's high confidence in predicting interaction regions.

\section{Discussion}
\textbf{Conclusion.} We plan to anticipate interaction intentions, encompassing the joint prediction of future hand trajectories and interaction hotspots with objects. 
We propose a novel \textbf{BOT} framework to establish an inherent connection between hand trajectories and contact points by introducing a \textbf{Bi-Progressive} mechanism, which eliminates the effect of error accumulation and achieves more accurate intention anticipation. Experimental results on three challenging benchmarks achieve state-of-the-art performance, thereby proving the superiority of our method in intention anticipation.

\textbf{Future Work.} \textbf{BOT} currently predicts hand trajectories and interaction hotspots on 2D images. However, human-object interactions occur in 3D space. To enable the model to be applied to real-world scenarios, we will consider the geometric structure and spatial relationships between objects in the scene in the future, providing the model with the ability to anticipate interaction intentions for the real world.

\par\vfill\par

\clearpage  

%
%
\bibliographystyle{splncs04}
\bibliography{main}
\end{document}